\documentclass[10pt,twocolumn,letterpaper]{article}

\usepackage{3dv}
\usepackage{times}
\usepackage{epsfig}
\usepackage{graphicx}
\usepackage{amsmath}
\usepackage{amssymb}
\usepackage{tabularx}

\usepackage{textcomp}
\usepackage{float}
\usepackage{dblfloatfix} 
\usepackage{multicol}

\usepackage{caption}
\usepackage{subcaption}
\usepackage{bbold}

\usepackage{ dsfont }

\newcommand{\mypara}[1]{\vspace{-2mm}\paragraph{#1}}

\newcommand{\X}{\mathcal{X}}
\newcommand{\Y}{\mathcal{Y}}
\newcommand{\F}{\mathcal{F}}

\newenvironment{Figure}
  {\par\medskip\noindent\minipage{\linewidth}}
  {\endminipage\par\medskip}
  
\newenvironment{Table}
  {\par\bigskip\noindent\minipage{\columnwidth}\centering}
  {\endminipage\par\bigskip}

\usepackage[pagebackref=true,breaklinks=true,colorlinks,bookmarks=false]{hyperref}

\DeclareMathOperator*{\argmin}{arg\,min}

\usepackage{geometrycollective}

\threedvfinalcopy 

\def\threedvPaperID{35} 

\ifthreedvfinal\fi
\begin{document}

\title{DPFM: Deep Partial Functional Maps}

\author{Souhaib Attaiki \qquad Gautam Pai  \qquad Maks Ovsjanikov\\
LIX, École Polytechnique, IP Paris\\
}

\maketitle
\thispagestyle{empty}

\global\csname @topnum\endcsname 0
\global\csname @botnum\endcsname 0

\begin{abstract}
We consider the problem of computing dense correspondences between non-rigid shapes with potentially significant partiality. Existing formulations tackle this problem through heavy manifold optimization in the spectral domain, given hand-crafted shape descriptors. In this paper, we propose the first learning method aimed directly at partial non-rigid shape correspondence. Our approach uses the functional map framework, can be trained in a supervised or unsupervised manner, and learns descriptors directly from the data, thus both improving robustness and accuracy in challenging cases. Furthermore, unlike existing techniques, our method is also applicable to partial-to-partial non-rigid matching, in which the common regions on both shapes are unknown a priori. We demonstrate that the resulting method is data-efficient, and achieves state-of-the-art results on several benchmark datasets.  Our code and data can be found online: \url{https://github.com/pvnieo/DPFM}.
\end{abstract}

\section{Introduction}

\begin{figure}
\begin{center}
\hspace{-0cm}
\includegraphics[width=1\linewidth]{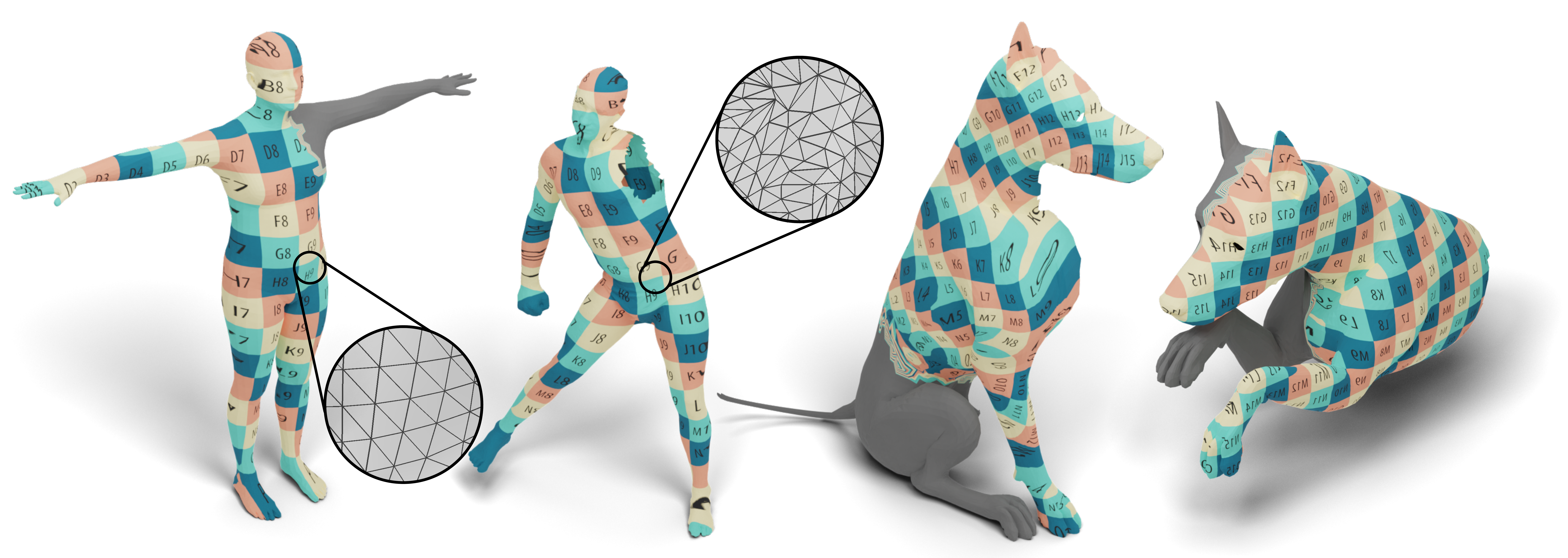}
\end{center}
\caption{Our method is highly effective for non-rigid partial shape matching, even in the challenging cases of significant mesh and sampling variability (left), and partial-to-partial matching (right). Our method detects both the overlapping regions between the shapes (non-common regions are colored in grey) and produces dense pointwise correspondences, visualized via texture transfer.}
\label{fig:teaser}
\end{figure}

Non-rigid shape correspondence is an essential problem in 3D Computer Vision, which enables a range of downstream tasks, such as statistical shape analysis \cite{pishchulin2017building} and deformation transfer \cite{sumner2004deformation} among many others.

A particularly challenging setting for this problem is computing correspondences between shapes in 3D that undergo both non-rigid deformations and exhibit strong partiality, such as missing parts or having holes, due, e.g., to acquisition errors. Several works have aimed to address this problem, in particular by adapting spectral techniques, within the functional map framework \cite{Rodol2016,Litany2017,Arbel2019,Wu2020,Xiang_2021_CVPR}. Spectral methods are inherently invariant under near-isometric deformations, making them attractive for non-rigid shape matching problems. Furthermore, objective functions promoting structural properties of functional maps have been designed through careful analysis of the alignment of Laplacian basis in the partial setting, e.g., \cite{Rodol2016,Litany2017}. As a result, these approaches lead to moderately-sized optimization problems and can achieve accurate results even under significant pose changes.

At the same time, most existing methods strongly rely on hand-crafted input features, and especially the popular local SHOT descriptors \cite{salti2014shot}. Unfortunately, these descriptors are known to be very sensitive to local mesh structure and can fail when comparing two shapes that have different sampling or connectivity \cite{Donati2020}. To remedy this issue, recent approaches have aimed at \emph{learning} features for shape correspondence in both supervised and unsupervised settings, e.g., \cite{masci2015geodesic,litany2017deep,halimi2019unsupervised}.

Despite this progress, however, existing learning-based methods are only geared towards computing correspondences between \emph{complete} non-rigid shapes, and can easily fail in the presence of strong partiality. This is because special care must be taken to account for partiality during both training and testing, as most losses and feature extraction networks are typically not applicable. 

In this paper, we present the first learning-based approach specifically geared towards finding correspondences between non-rigid shapes under strong partiality. For this, we design a novel architecture that can accommodate both partial-to-full and partial-to-partial correspondence. Our method can be trained in a supervised and unsupervised manner. We use the functional map framework but learn features from data, making our method robust both to change of poses and to changes in mesh sampling and connectivity (see Figure \ref{fig:teaser}).

\vspace{2mm}

Our contributions can be summarized as follows:
 \begin{itemize}
   \setlength\itemsep{0.1em}
     \item We propose a new deep learning method for non-rigid partial shape correspondence, which is robust under mesh changes,  predicts both the correspondences \emph{and the region of overlap}, and is the first learning-based approach for partial-to-partial non-rigid matching.
    \item We provide theoretical analysis of the limitations of a Siamese architecture for partial functional maps, and develop a module that enables communication between features.
     \item We study the robustness of existing methods and introduce two new datasets, both for partial-to-partial and partial-to-full matching with significant mesh variability. Our method achieves state-of-the-art on both standard and these new benchmarks.
 \end{itemize}

\section{Related Work}
Shape matching is a vast and well-studied area of Computer Vision. Below we review works most closely related to ours, focusing especially on non-rigid partial shape correspondence, and refer the interested readers to recent surveys \cite{van2011survey,biasotti2016recent,sahilliouglu2020recent} for a more in-depth treatment of this field.

\mypara{Partial rigid correspondence}
Partial matching has been particularly well-studied in the context of \emph{rigid} shape alignment, motivated especially by shape reconstruction problems from partial scans \cite{castellani20203d,bellekens2014survey}. While early methods aimed to address this problem primarily using the classical ICP and its robust variants, e.g., \cite{rusinkiewicz2001efficient,chetverikov2002trimmed,du2011robust,bouaziz2013sparse}, more recent approaches focus on learning robust features that can be used for rigid alignment,  \cite{zeng20173dmatch,deng2018ppfnet,choy2019fully,gojcic2019perfect}. We note that several recent approaches in the latter category aim to directly address partiality, typically through outlier filtering, keypoint detection, or even explicit overlap prediction \cite{wang2019prnet,gojcic2019perfect,pais20203dregnet,huang2021predator}. While these methods achieve impressive performance even in the presence of extreme partiality \cite{huang2021predator}, they are all limited to rigid alignment, which has a small number of degrees of freedom -- indeed, only three point pair correspondences are sufficient to recover a rigid transformation.


\mypara{Learning for full-to-full non-rigid correspondence}
Learning strategies have also been applied to address non-rigid shape correspondence problems, especially pioneered by techniques  in geometric deep learning \cite{bronstein2017geometric,cao2020comprehensive}. Most approaches in this field treat shape correspondence as a vertex labeling problem \cite{monti2017geometric,boscaini2016,masci2015geodesic,poulenard2018multi,wiersma2020cnns}, thus mapping each shape to some predefined canonical template. In contrast, methods based on the functional map representation have recently allowed to predict and penalize \emph{entire maps} between shape pairs \cite{litany2017deep,halimi2019unsupervised,roufosse2019unsupervised,Donati2020} achieving better accuracy and generalization power. While approaches in the former category have been applied in the context of partial-to-full matching in, e.g., \cite{boscaini2016} and \cite{Li_2020_CVPR}, they are unfortunately limited to matching shapes to a predefined template and furthermore have been observed to not be robust under sampling and connectivity changes \cite{sharp2021diffusionnet}. On the other hand, existing learning functional maps-based techniques treat the correspondence problem globally but are geared towards complete shape matching. Some techniques can generalize to unseen partial matching \cite{SharmaO20} but do not outperform axiomatic methods, in part, as they are not explicitly trained to deal with partiality.

\mypara{Spectral methods for partial non-rigid matching}
In the realm of axiomatic spectral techniques and optimization methods, the partial functional map framework (PFM) \cite{Rodol2016} is one of the important methods to tackle partial shape correspondence. The PFM scheme which is principally designed to tackle a part-to-full correspondence problem alternates between estimating a functional map and a part-region indicator function on the full shape, where both  terms are strongly regularized. 
Follow-up works like \cite{Litany2016,Cosmo2016} extended the PFM approach to both cluttered non-rigid correspondence and multi-part settings where potentially multiple functional maps are optimized simultaneously.

Litany and colleagues \cite{Litany2017} proposed a fully spectral approach for partial shape correspondence which avoids region-based optimization. Similar in spirit to joint diagonalization schemes like \cite{kovnatsky2013coupled},  this method employs a manifold optimization scheme to estimate a linear transformation to align the spectral basis between the shapes and is also posed to handle part-to-part shape correspondence.

More recently, iterative spectral methods \cite{Melzi2019} have been shown to effectively handle the partial setting, where the slanted structure of the functional map is first estimated using the same spectral relation as proposed in PFM and then used in an upsampling strategy to refine an initial correspondence. Similarly, \cite{Xiang_2021_CVPR} and \cite{Wu2020} use refinement via spectral upsampling by combining an array of objectives promoting structural properties of the computed functional maps, while \cite{Arbel2019} target partial shape matching using properties of the Nearest-Neighbor Field (NNF). Finally, \cite{rampini2019correspondence} use a correspondence-free method for localizing the region on a full shape in the context of partial-to-full shape comparison.

We highlight that all of the existing axiomatic techniques strongly rely on input descriptors either for initialization or even for map optimization. Practical implementations typically employ SHOT descriptors within partial non-rigid matching pipelines. However, such handcrafted descriptors are known to be sensitive to the discretization of the surface, which severely affects the robustness of all resulting methods. Instead, we put special emphasis on the robustness of our approach and show that it can produce accurate results even under significant mesh changes.

\mypara{Partial-to-partial} Most existing methods assume that there exists a corresponding point on the target (full) shape for every point on the source (partial) shape. To address the more general partial-to-partial matching, FSP \cite{Litany2017} proposed a formulation for this setting, but without providing any practical evaluation. 

Similar in spirit to \cite{Rodol2016}, \cite{Cosmo2016} demonstrates a partial-to-partial matching pipeline between a source (model) and a target shape (scene) - that potentially comprises an overlapped part of the model and other non-related cluttered objects. 
However, this method is highly dependent on an initialization procedure like \cite{rodola2012game} which ultimately depends on the quality of robust input descriptors, unlike SHOT which is usually chosen by default. 

Finally, Litany and colleagues  \cite{Litany2016} proposed a solution to the non-rigid puzzle problem, where partial-to-full matching was extended to include possible additional clutter. However, their solution is constrained by assumptions on the existence of missing parts and ultimately still relies on a partial-to-full matching component.

In contrast to all of these existing techniques, our method completely avoids the need for hand-crafted input descriptors and allows to learn features directly from the 3D data. Furthermore, our functional map-based learning pipeline is geared towards partiality and can learn from limited training data, due to the strong spectral regularization. Thus, our method is virtually parameter-free and strongly outperforms existing methods, especially in the more realistic setting of shapes with significantly different mesh structures.


\section{Motivation and Method Overview}
As mentioned above, our main goal is to develop a learning-based method for partial non-rigid shape correspondence, while avoiding the use of hand-crafted features. For this, we follow the general strategy of recent learning-based spectral techniques \cite{litany2017deep,halimi2019unsupervised,SharmaO20,Donati2020}, and especially Deep Geometric Functional Maps introduced in \cite{Donati2020}. This approach can be summarized as follows:
\begin{enumerate}
\item Given the source and target shapes $\mathcal{X}, \mathcal{Y}$, a trainable Siamese feature network $\mathcal{F}_{\Theta}$ produces $m$ feature functions $\mathcal{F}_{\Theta}(\mathcal{X}) = \{f_{1}, f_{2}, ..., f_{m}\}$ and $\mathcal{F}_{\Theta}(\mathcal{Y}) = \{g_{1}, g_{2}, ..., g_{m}\}$.
\item These features are projected onto the Laplace-Beltrami basis to obtain their coefficients $\mathbf{a}_i = \Phi_{\mathcal{X}}^{+} f_i$ and $\mathbf{b}_i = \Phi_{\mathcal{Y}}^{+} g_i$, where $\Phi_{\mathcal{X}},\Phi_{\mathcal{Y}}$ are the matrices that store, as columns, the pre-computed set of Laplace-Beltrami eigenfunctions of shapes $\mathcal{X}$ and $\mathcal{Y}$ respectively, while $+$ denotes the Moore-Penrose pseudo-inverse.
\item Finally, $\{\mathbf{a}_i, \mathbf{b}_i\}$ are stacked as columns of matrices $\mathbf{A},\mathbf{B}$ and fed to a regularized functional map layer, which uses a linear system to compute the functional map \cite{ovsjanikov2012functional} in a reduced basis:
\begin{align}
\label{eq:fmap_energy}
\hspace{-3mm}C_{opt} = \argmin_{C} \| C\mathbf{A} - \mathbf{B} \| + \lambda \|C \Delta_{\mathcal{X}} - \Delta_{\mathcal{Y}} C \|.
\end{align}
Here $\Delta_{\mathcal{X}},\Delta_{\mathcal{Y}}$ are diagonal matrices of Laplace-Beltrami eigenvalues of the corresponding shapes and $\lambda$ is a scalar hyper-parameter.
\item  The computed functional map $C_{opt}$ is compared to the ground truth $C_{gt}$ one by imposing the Frobenius norm loss $\mathcal{L}(C_{opt}) = \| C_{gt} - C_{opt}\|_F^2$, and this loss is used to train the parameters $\Theta$ of the feature network $\mathcal{F}_{\Theta}$ in a supervised manner.
\end{enumerate}
As mentioned in that work and several recent follow-ups \cite{Donati2020,SharmaO20,sharp2021diffusionnet}, using the reduced Laplacian basis helps to regularize feature learning and allows to learn even from a limited set of training shapes, while avoiding the use of a predefined template as done in works that treat shape correspondence as a vertex labeling problem, e.g., \cite{boscaini2016,monti2017geometric,wiersma2020cnns}.

\begin{figure*}
\includegraphics[width=1\linewidth]{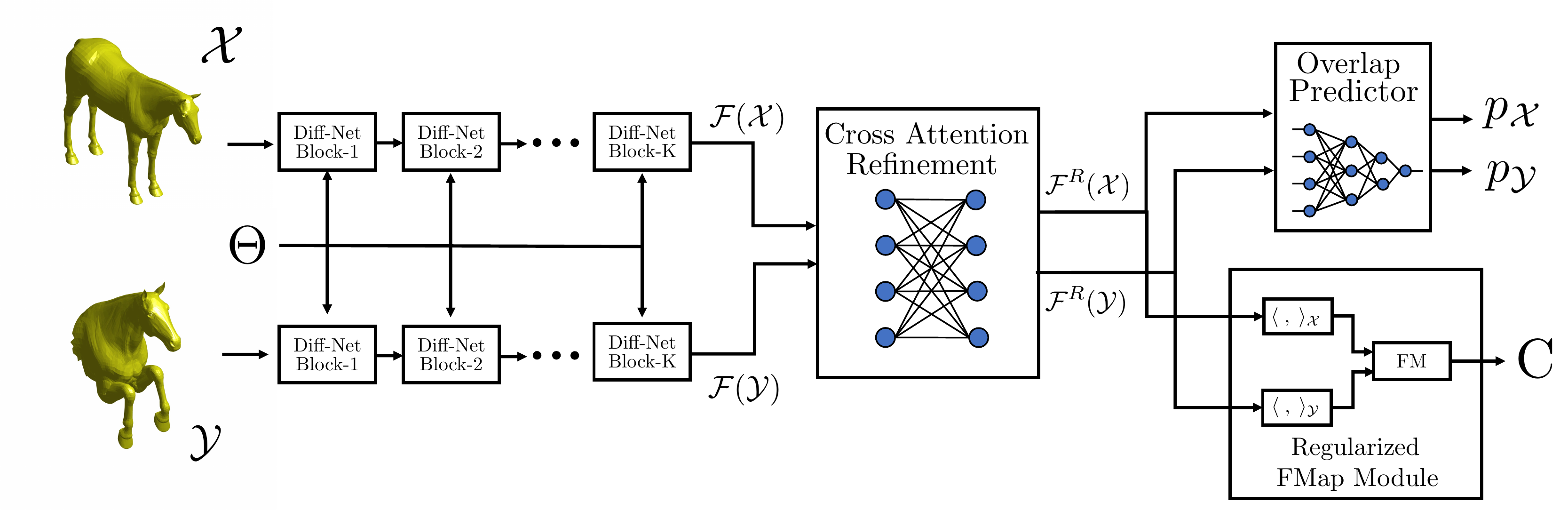}
\vspace{-6mm}
\caption{Overview of our approach. Raw input geometry of a pair of shapes is processed by a Siamese network with shared weights resulting in features $\mathcal{F}({\mathcal{X}})$ and $\mathcal{F}({\mathcal{Y}})$. These are then refined using a Cross Attention Refinement module allowing the features on the two shapes to communicate. The refined features are then used, first, to predict the overlap region using the Overlap Predictor Module, and second, to compute the functional map using the parameter-free Regularized FMap module.}
\label{fig:overview}
\end{figure*}

\mypara{Overview} Our approach follows the same general approach as in \cite{Donati2020}, while making several key contributions (See Figure \ref{fig:overview} for the overview).

First, we replace the Siamese network $\mathcal{F}_{\Theta}$ with a combination of a feature extractor and a \emph{Cross Attention refinement} block. Our main motivation is that in the context of partial correspondence, the feature extractor on each shape should be made aware of the other shape. In the functional map language, if $\mathbf{a}$ and $\mathbf{b}$ are coefficients of the feature functions on the source and target respectively, and if $\mathbf{a}=0$ because, e.g., a particular part is missing then $C\mathbf{a} = \mathbf{b}$ will only hold if $\mathbf{b}=0$. However, a Siamese network is not aware of the descriptors on the source shape and would produce a non-zero feature $\mathbf{b}$ that characterizes the existing part on the target shape. We provide an in-depth theoretical analysis of this effect, and highlight the limitations of a Siamese architecture for partial functional maps in the supplementary.

Motivated by these considerations we introduce the Attention-based Feature refiner (Section \ref{subsec:refiner}) that takes as input features on both shapes and produces refined features of each shape that are ``informed'' by the features on the other shape. We remark that since functional maps operate \emph{in the opposite direction} of point-to-point ones, the analysis above applies when computing a pointwise map from  either a full or a partial to another partial shap. However, we have found that the ``communication'' between features is generally useful across a range of matching settings.

Our second major contribution is that in addition to the functional map, we also train an Overlap predictor (Section \ref{subsec:overlap}) that produces the region on each shape that is expected to exist on the other shape. This helps us to both produce a dense pointwise map and the region in which it is expected to be accurate thus allowing us to deal with full-to-partial and partial-to-partial non-rigid matching.

Finally, we make several other architecture modifications that improve its accuracy and robustness: we use more powerful intrinsic and sparse feature extractors (Section \ref{subsec:feat}) and a regularized functional map layer adapted to the partial setting (Section \ref{subsec:fmap_layer}). Below we provide the details of each of these building blocks.


%
%
%
%

\subsection{Feature extractor}
\label{subsec:feat}
As mentioned above, and as shown in Figure \ref{fig:overview}, the first block in our pipeline is a feature extractor module.

The goal of this module is to take as input the raw shape geometry and produce meaningful descriptors that will be used downstream for spectral matching. This module should be robust to rigid and non-rigid deformations, to the surface sampling, and more importantly, to the challenges introduced by partiality. Throughout our work, we use the recent DiffusionNet architecture \cite{sharp2021diffusionnet} that is based on intrinsic operations for information diffusion on the surface, and has been shown to be robust under shape discretization changes. In cases of extreme partiality such as the Holes setting of the SHREC 2016 partiality benchmark \cite{cosmo2016shrec}, we also experimented with SparseConvNet \cite{graham20183d}, which can better handle very sparse data. This leads to a variant of our approach that we call DPFM$_{\text{sparse}}$. Unless specified otherwise, DPFM always denotes our approach with the DiffusionNet feature extractor.
%

Following  \cite{Donati2020}, our feature extractor is applied in a Siamese way, i.e the same network with identical weights is applied to the source $\mathcal{X}$ and target $\mathcal{Y}$ shapes. This network produces an $m$-dimensional vector per point, or equivalently $m$ feature functions on the surface, that we denote $\mathcal{F}({\mathcal{X}})$ and $\mathcal{F}({\mathcal{Y}})$ respectively. We observe that the surface-aware feature extractors like DiffusionNet and SparseConv typically outperform the point-based KPConv \cite{thomas2019kpconv} used in \cite{Donati2020}.

\subsection{Attention-based Feature Refiner}
\label{subsec:refiner}
The features produced by the feature extractors encode the geometry and context of each shape independently and have no knowledge of the other shape, making predicting the overlap region impossible. We remedy this by letting the features \emph{communicate} using a cross attention block, using a design inspired by the recent Predator network \cite{huang2021predator} architecture, that was introduced for partial \textit{rigid} alignment.


Specifically, we construct a bipartite graph $(\mathcal{V}, \mathcal{E})$, where every point on shape $\mathcal{X}$ is connected to all points on shape $\mathcal{Y}$, and we associate to each node the point-wise features learned from the feature extractor. The message passing formulation of \cite{kipf2017semi} is used to propagate information through this graph.

Following the Transformer architecture \cite{attentionneed}, for each node $i$, a representation query $q_i$ is learned, and it retrieves learned values $v_j$, based on their learned attributes, the keys $k_j$. The final message that is passed to node $i$ has the form:
$$m_{\mathcal{E} \rightarrow i} = \sum_{j, (i, j) \in \mathcal{E}} \alpha_{ij} v_j$$
where $\alpha_{ij}$ are the attention weights, based on the key-query similarity $\alpha_{ij} = \text{Softmax}_j (q_i^{\top} k_j / \sqrt{d})$.

The queries $q_i \in \mathbb{R}^d$, keys $k_i \in \mathbb{R}^d$ and values $v_i \in \mathbb{R}^d$ are learned as a linear projection of learned features of the previous block, using  a learnable linear layer.

Finally, the value of the node $i$ is updated using the formula:
$$x_i = x_i + MLP([x_i || m_{\mathcal{E} \rightarrow i}])$$
where $[. ||.]$ stands for concatenation, and $MLP$ is a three layer perceptron \cite{MLP} with ReLU activations \cite{xu2015empirical} and instance normalization \cite{ulyanov2016instance}.

In practice and in line with existing literature \cite{attention2018graph}, we improve the expressivity of the model using multi-head attention \cite{attentionneed} with four heads. Also, in order to reduce the computation and memory footprint of this block, we sub-sample the two shapes using the farthest point sampling FPS \cite{pointnet++} before constructing the bipartite graph. The learned features are then linearly interpolated to the missing points.

The feature refiner module takes as input features $\mathcal{F}(\mathcal{X}),\mathcal{F}(\mathcal{Y})$ and produces \textit{refined} features that we denote $\mathcal{F}^R(\mathcal{X}),\mathcal{F}^R(\mathcal{Y})$.

\begin{figure}[t]
\begin{center}
   \includegraphics[width=1\linewidth]{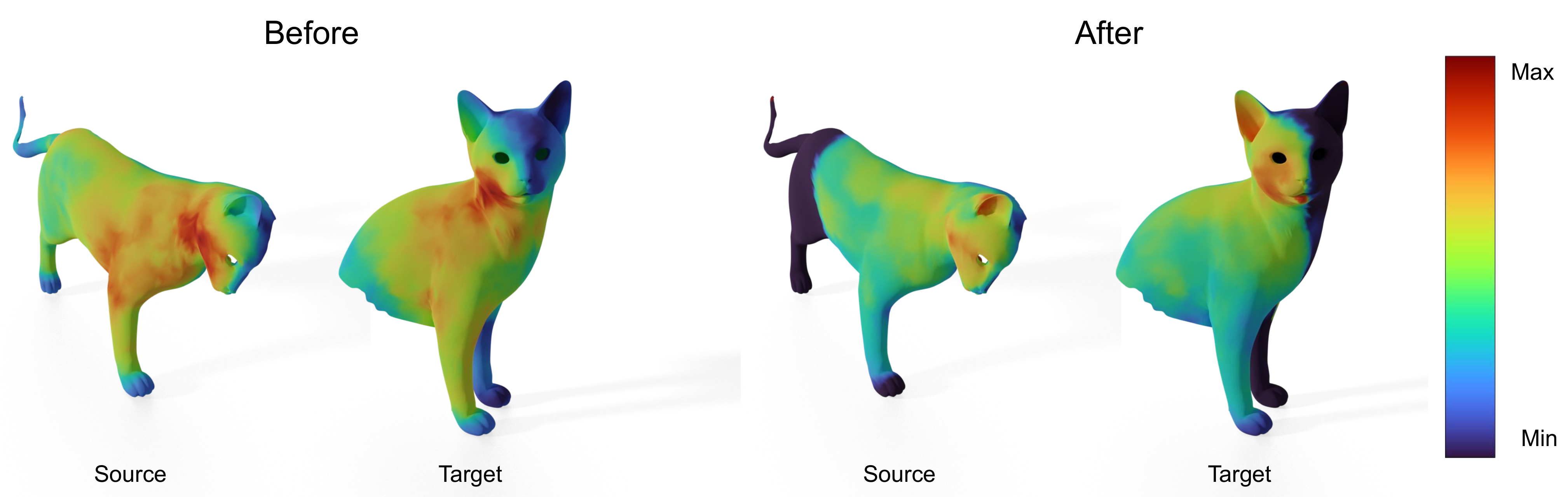}
\end{center}
\vspace{-1.8em}
\caption{Visualizing the per-point feature magnitude, before and after the Cross-Attention Refinement module.}
\label{fig:intensity_feature}
\vspace{-1.5em}
\end{figure}

As mentioned above, our Cross-Attention Refinement module enables the \textit{communication} between features on the two shapes and thus allows the features on the overlap region to synchronize, while down-weighing the features outside the overlap.  We illustrate this effect in Figure \ref{fig:intensity_feature}, where we show the intensity of the learned features, before and after the refinement module. Note that before refinement, the features are uniformly distributed, but after the refinement, the features on the missing region tend to zero, as suggested by our theoretical analysis.

Additional analysis and quantitative evaluation of the Cross-Attention Refinement module are provided in the supplementary materials.

\subsection{Overlap Predictor}
\label{subsec:overlap}
The combination of the two previous blocks produces features that capture geometric properties of each shape and are conditioned by the distribution of features on the other shape. 

We then use these features to predict the overlap probability $p_{\mathcal{X}} \in \mathbb{R}^{n_{\mathcal{X}}}$ and $p_{\mathcal{Y}} \in \mathbb{R}^{n_{\mathcal{Y}}}$ where ${n_{\mathcal{X}}}$ and ${n_{\mathcal{Y}}}$ is the number of points on shape $\mathcal{X}$ and $\mathcal{Y}$ respectively. For this we use a simple two layer MLP with ReLU activations, and a Sigmoid layer \cite{nwankpa2018activation} at the end, applied independently to each point.  We interpret, e.g., $p_{\mathcal{X}}(i)$ as the probability of vertex $i$ of shape $\mathcal{X}$ to exist on shape $\mathcal{Y}$.

\subsection{Regularized FMap Layer}
\label{subsec:fmap_layer}
While the previous block predicts the \textit{existence} of points across shapes, our main goal is to establish dense point-to-point correspondences between co-existing points. To this end we use a regularized functional map module that aims to predict the functional map between shape $\mathcal{X}$ and $\mathcal{Y}$, using the refined features $\mathcal{F}^R_{\mathcal{X}}$ and $\mathcal{F}^R({\mathcal{Y}})$. We therefore project the feature functions onto their respective spectral basis consisting of the first $k$ Laplace-Beltrami eigenfunctions: $\mathbf{A} = \Phi_{\mathcal{X}}^{+} \mathcal{F}^R({\mathcal{X}})$ and $\mathbf{B} = \Phi_{\mathcal{Y}}^{\dagger} \mathcal{F}^R({\mathcal{Y}})$. 

We then use these refined coefficient matrices to solve for the optimal functional map. Rather than optimizing the energy defined in \eqref{fmap_energy} above, as done in \cite{Donati2020}, we modify the regularizer for improved robustness.

Specifically, we first observe that the commutativity term $E_{comm}(C) =  ||C\Delta_{\mathcal{X}} - \Delta_{\mathcal{Y}}C||_F^2$ can be written as a \emph{mask} applied on the functional map $E_{comm}(C) = \sum_{i, j} M_{ij} C_{ij}^2$, where $M_{ij} =  (\Lambda_{\mathcal{Y}}(i) - \Lambda_{\mathcal{X}}(j))^2$ (see Eq. (3) in \cite{Ren2019}). Then, we note that the closed form minimization technique used in \cite{Donati2020} to make the computation of the optimal functional map is differentiable  and can be applied to \emph{any} mask matrix $M$ regardless of whether it is associated with the Laplacian commutativity term or not. 

This motivates our objective to choose a mask that is well-suited for the partial setting. In our experiments, we test the original mask formulation, dubbed the Laplacian mask, the slanted mask proposed in \cite{Rodol2016}, and the resolvent mask proposed in \cite{Ren2019}. As argued in \cite{Ren2019}, the resolvent mask has a more solid theoretical foundation. In our work, we also observed that the resolvent mask consistently leads to better results in practice, and we adopt it throughout all our experiments.

In summary, we compute the functional map by minimizing the following energy:
\begin{align}
    C_{opt} = \argmin_{C} \|C \mathbf{A} - \mathbf{B}\|_F^2 + \lambda \sum_{ij} C_{ij}^2 M_{ij},
    \vspace{-3mm}
\end{align}
where 
%
%
\begin{equation*}
\resizebox{0.48\textwidth}{!}{$M_{ij} = \left( \frac{\Lambda_{\mathcal{Y}}(i)^{\gamma}}{\Lambda_{\mathcal{Y}}(i)^{2\gamma} + 1} - \frac{\Lambda_{\mathcal{X}}(j)^{\gamma}}{\Lambda_{\mathcal{X}}(j)^{2\gamma} + 1} \right)^2 + \left( \frac{1}{\Lambda_{\mathcal{Y}}(i)^{2\gamma} + 1} - \frac{1}{\Lambda_{\mathcal{X}}(j)^{2\gamma} + 1} \right)^2$}
\end{equation*}
and $\gamma$ is the resolvent Laplacian parameter, which we will take equal to 0.5 in all our experiments.






\subsection{Training losses}
As all of our architecture blocks are differentiable, we train the network end to end using ground truth correspondences using backpropagation. Our overall loss is composed of three main components:

\paragraph{Spectral Loss} given a ground truth point to point map, we convert it to a functional map, and penalize the deviation of our predicted map from it using the Frobenius norm $\mathcal{L}_{spec}(C_{opt}) = \| C_{gt} - C_{opt}\|_F^2$ as done in \cite{Donati2020}.
The ground truth map is computed as follows:
$C_{gt} = \Phi_{\mathcal{Y}}^{+} \Pi_{\mathcal{X}\mathcal{Y}} \Phi_{\mathcal{X}}$
where $\Pi_{\mathcal{X}\mathcal{Y}}$ is the ground truth point to point matrix. I.e., $\Pi_{\mathcal{X}\mathcal{Y}}(i,j) = 1$ if and only if vertex $i$ on $\mathcal{X}$ is matched to the vertex $j$ on $\mathcal{Y}$ and 0 otherwise. Note that, unlike the full shape matching, in the case of partial correspondence, $\Pi_{\mathcal{X}\mathcal{Y}}$ might have entire rows that are zero, associated with vertices for which no match exists on the target.

\paragraph{PointInfoNCE Loss} In our experiments, we observed that the spectral loss alone is not sufficient to train the feature extractor, especially in the challenging case of holes, where the ground truth functional map can fail to provide important high-frequency information. To remedy this, we use a contrastive loss, where the distance between extracted features of matched points should be minimized, meanwhile, the distance between unmatched points features should be maximized. In our experiments, we used PointInfoNCE Loss \cite{Xie_2020} which can be expressed as:
%
\begin{equation*}
\vspace{-2mm}
\resizebox{0.48\textwidth}{!}{$\mathcal{L}_{nce}(F_{\mathcal{X}}^R, F_{\mathcal{Y}}^R) = -\hspace{-2mm}\sum_{(i,j)\in \mathcal{P}} \log \frac{\exp(F_{\mathcal{X}}^R(i) \cdot F_{\mathcal{Y}}^R(j)/ \tau) }{\sum_{(.,k)\in \mathcal{P}}\exp(F_{\mathcal{X}}^R(i) \cdot F_{\mathcal{Y}}^R(k)/ \tau) }$}
\end{equation*}

Here $\mathcal{P}$ is the set of matched points, computed using ground truth point to point map, $\tau$ is a scaling parameter, and $F_{\mathcal{X}}^R(i)$ is the refined feature of point $i$ in shape $\mathcal{X}$. Note that we use $F_{\mathcal{X}}^R$ to denote the refined features $\mathcal{F}^R(\mathcal{X})$ for compactness.

\paragraph{Overlap Loss} To train the overlap predictor module, we penalize the computed probabilities using a binary classification loss. The ground truth labels $y_{\mathcal{X}}$ and $y_{\mathcal{Y}}$ are extracted from the ground truth point to point map: $$\mathcal{L}_{over}(p_{\mathcal{X}}, p_{\mathcal{Y}}) = \frac{1}{2} (\mathcal{L}_{classif}(p_{\mathcal{X}}, y_{\mathcal{X}}) + \mathcal{L}_{classif}( p_{\mathcal{Y}}, y_{\mathcal{Y}}))$$
where $\mathcal{L}_{classif}$ is the binary cross entropy loss.
$$\mathcal{L}_{classif}(x, y) = \frac{1}{|x|} \sum_{i=1}^{|x|} y_i \log(x_i) + (1-y_i)\log(1 - x_i)$$

\paragraph{Total loss} our final loss is just a weighted sum to the previous losses:
$\mathcal{L} = \lambda_1 \mathcal{L}_{spec} + \lambda_2 \mathcal{L}_{nce} + \lambda_3 \mathcal{L}_{over}$.

We use the same values of all of our hyperparameters throughout all of our experiments. We provide the complete implementation details and hyperparameter values in the supplementary. 

\vspace{-2mm}\paragraph{Unsupervised variant} While these losses depend on the presence of a ground truth correspondence, our pipeline can also be adapted to an \emph{unsupervised} setting by using losses that promote desirable structural properties of the functional maps. We provide the details and evaluate one such possibility in the supplementary materials.

\mypara{Ablation Study}
Our method comprises multiple building blocks that we consider essential for optimal performance. In the supplementary, we report an ablation study that demonstrates the efficacy of individual components. More significantly, we observe that the spectral loss is crucial for the convergence of the overlap predictor network and also report that the \emph{resolvent} mask is best suited in our setting. 
\section{Experiments and Analysis}
\label{sec:experiments}
In this section, we provide extensive experiments highlighting the accuracy and robustness of our method on a range of non-rigid partial shape matching tasks. We consider three main challenges: partial-to-full, partial-to-partial, and finally partial-to-full under extreme variability in mesh discretization. 


\begin{figure*}[h]
\begin{subfigure}{.5\textwidth}
  \centering
  \includegraphics[width=0.95\linewidth]{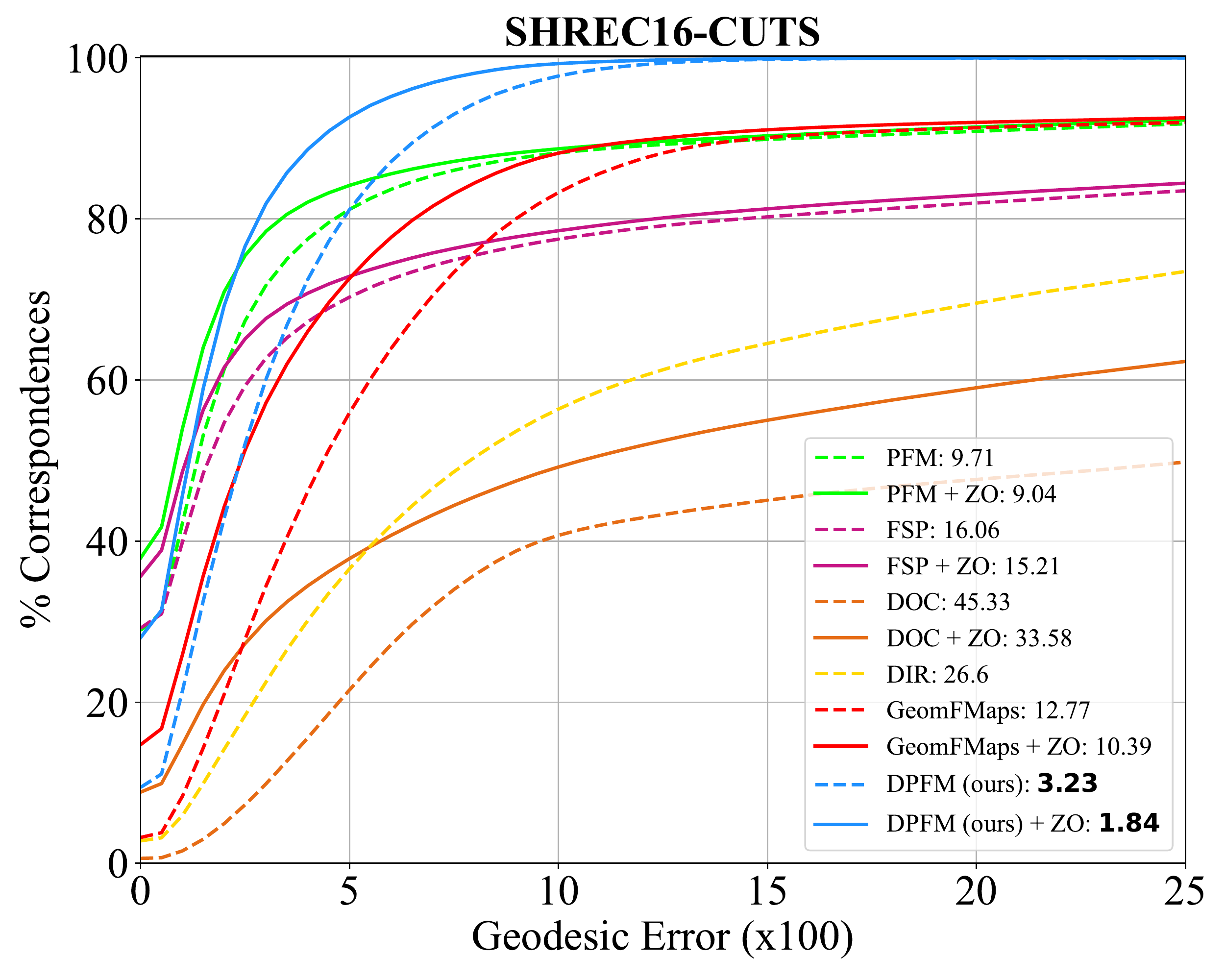}
  \label{fig:princeton_p2f_cuts}
\end{subfigure}%
\begin{subfigure}{.5\textwidth}
  \centering
  \includegraphics[width=0.95\linewidth]{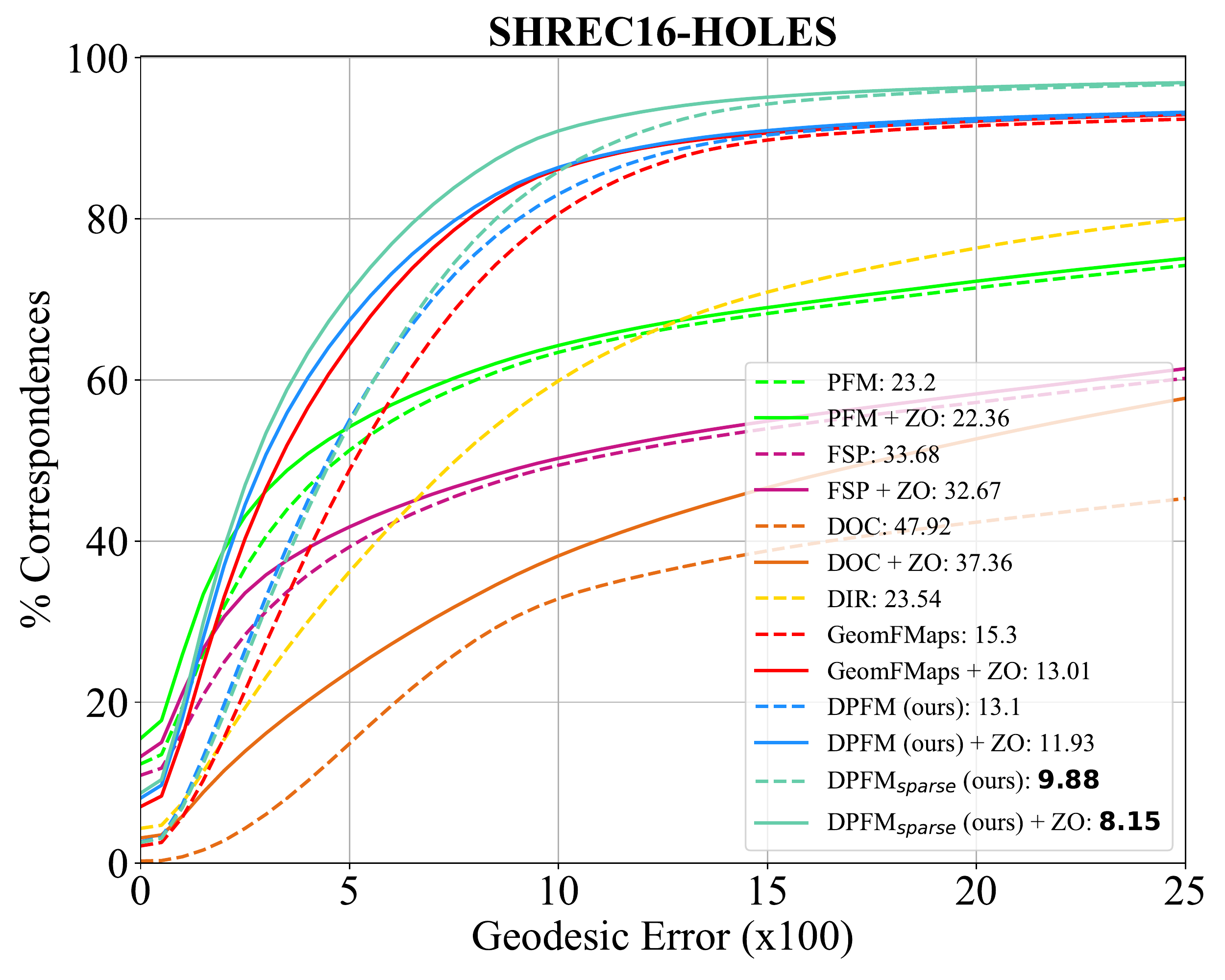}
  \label{fig:princeton_p2f_holes}
\end{subfigure}
\vspace{-1mm}
\caption{Correspondence quality of different methods on the test set of SHREC'16 Partial Benchmark, both on cuts (left) and holes (right). Our method outperforms all the competing methods and achieves state-of-the-art results. Mean errors of all methods are reported in the legend.\vspace{-4.5mm}}
\label{fig:princeton_p2f}
\end{figure*}

\subsection{Datasets}
\label{sec:datasets_prez}
We evaluate our method qualitatively and quantitatively using the following benchmarks. See the supplementary for visualization of shapes from each of the following datasets.

\textbf{The SHREC'16 Partial Correspondence Benchmark \cite{cosmo2016shrec}} is a widely used benchmark to evaluate partial non-rigid shape matching pipelines. This dataset consists of 76 nearly isometric shapes, divided into 8 classes (humans and animals). Each class has a 'null' shape to which the partial shapes will be matched. As an additional challenge, all shapes are individually re-meshed to 10K vertices. 

There are two subsets of \textbf{SHREC'16}, namely \textbf{CUTS} and \textbf{HOLES}. \textbf{CUTS} is constructed by cutting the shapes with a random plane at different orientations. The \textbf{HOLES} dataset is more challenging and contains shapes with surfaces that have been eroded by various random seeds, causing them to have actual holes and very irregular cuts.

\textbf{SHREC'16 Part-to-Part dataset (\textbf{CP2P})}: We introduce a new dataset to evaluate partial-to-partial shape matching. \textbf{CP2P} is based on the \textbf{CUTS} training set in  \cite{cosmo2016shrec} and contains the same 8 classes of humans and animals in different poses and partialities and establishes a total of 300 pairs for matching. The train and test sets are divided randomly with 80\% pairs for training, and the remaining 20\% for testing. We highlight that, similar to \textbf{SHREC'16} the ground truth map is only provided for shapes from the same class. The overlap between the source and target shapes varies from 10\% to 90\% of the total area.

\textbf{Partial FARM dataset (\textbf{PFARM})}: We introduce a new dataset with the objective of testing the robustness of partial shape matching methods to significantly varying mesh connectivity and sampling. Our dataset (\textbf{PFARM}) is designed using the SHREC'19 dataset \cite{shrec19connectivity} and is an extension of the recently introduced FARM partial dataset \cite{Kirgo2020}. \textbf{PFARM} comprises of 27 test pairs resulting from 28 human shapes (one null shape). The shapes have significantly different meshing, vertex density, undergo varied rigid and non-rigid deformations and have significant partiality. 

\subsection{Baselines}
We compare with the following prior partial shape matching methods: Partial Functional Maps (PFM) \cite{Rodol2016}, Fully Spectral Partial Shape Matching (FSP) \cite{Litany2017}, Matching Deformable Objects in Clutter (DOC) \cite{Cosmo2016} and also a recent method: Dual Iterative Refinement (DIR) \cite{Xiang_2021_CVPR}. Since our work is the first to tackle a \emph{learning} approach to non-rigid partial shape matching, we also include GeomFMaps \cite{Donati2020} as a \emph{learning} baseline, since that method is similar in spirit to our framework. In addition, we refine almost all methods with the partial variant of the Zoomout algorithm \cite{Melzi2019}. For all  competing methods, we use the original code released by the authors and apply the best parameters reported in the respective papers.

\mypara{Evaluation Protocol}
We follow the Princeton protocol \cite{Kim2011} for evaluating all correspondences, as used in all recent works. Specifically, we compute the pointwise geodesic error between the predicted maps and the ground truth map and normalize by the square root of the total surface area of the target shape.
\subsection{Part To Full Shape Matching}
\label{subsec:p2f_exp}
We first show the results of our method for partial-to-full matching on \textbf{SHREC'16}. 
We individually train our networks on the train set of \textbf{SHREC'16} comprising of 120 pairs (\textbf{CUTS}) and 80 pairs (\textbf{HOLES}) respectively. 


It should be noted that contrary to many of the previous works, which are evaluated on the train set of \textbf{CUTS} and \textbf{HOLES}, we follow the original \textbf{SHREC'16}  Benchmark protocol, and test our method, and all the baselines \emph{only on the test set}, which contains 200 pairs each for \textbf{CUTS} and \textbf{HOLES}. The test set contains unseen shapes during training and provides a diverse and \emph{significantly} more challenging set of partiality scenarios.

As reported in Figure \ref{fig:princeton_p2f}, our method achieves state-of-the-art results across all settings in \textbf{SHREC'16} outperforming both axiomatic and learning-based prior methods. Additionally, we show the performance of our sparse variant Ours$_{\text{sparse}}$ on the \textbf{HOLES} dataset, as the latter can handle particularly sparse data, and show that
our architecture is not limited to the choice of the feature extractor.

In the supplementary materials, we also report two additional results in this setup. First, we plot the mean geodesic error as a function of the percentage of partiality. We observe that our method achieves the lowest mean error among all competing methods, and the error is stable even with an increasing amount of partiality, demonstrating the robustness of our method. Secondly, we also report the quality of overlap region prediction (which is a feature of only our method and PFM) by plotting the intersection over union (IOU) of the predicted region and the ground truth. 
We observe that our method achieves a high IOU, outperforming PFM, especially for \textbf{CUTS}, which demonstrate the quality of the predictions made by the overlap module. 
\subsection{Robustness to Mesh Discretization}
\label{subsec:robust_p2f_exp}

\begin{figure}[]
\begin{center}
\includegraphics[width=0.92\linewidth]{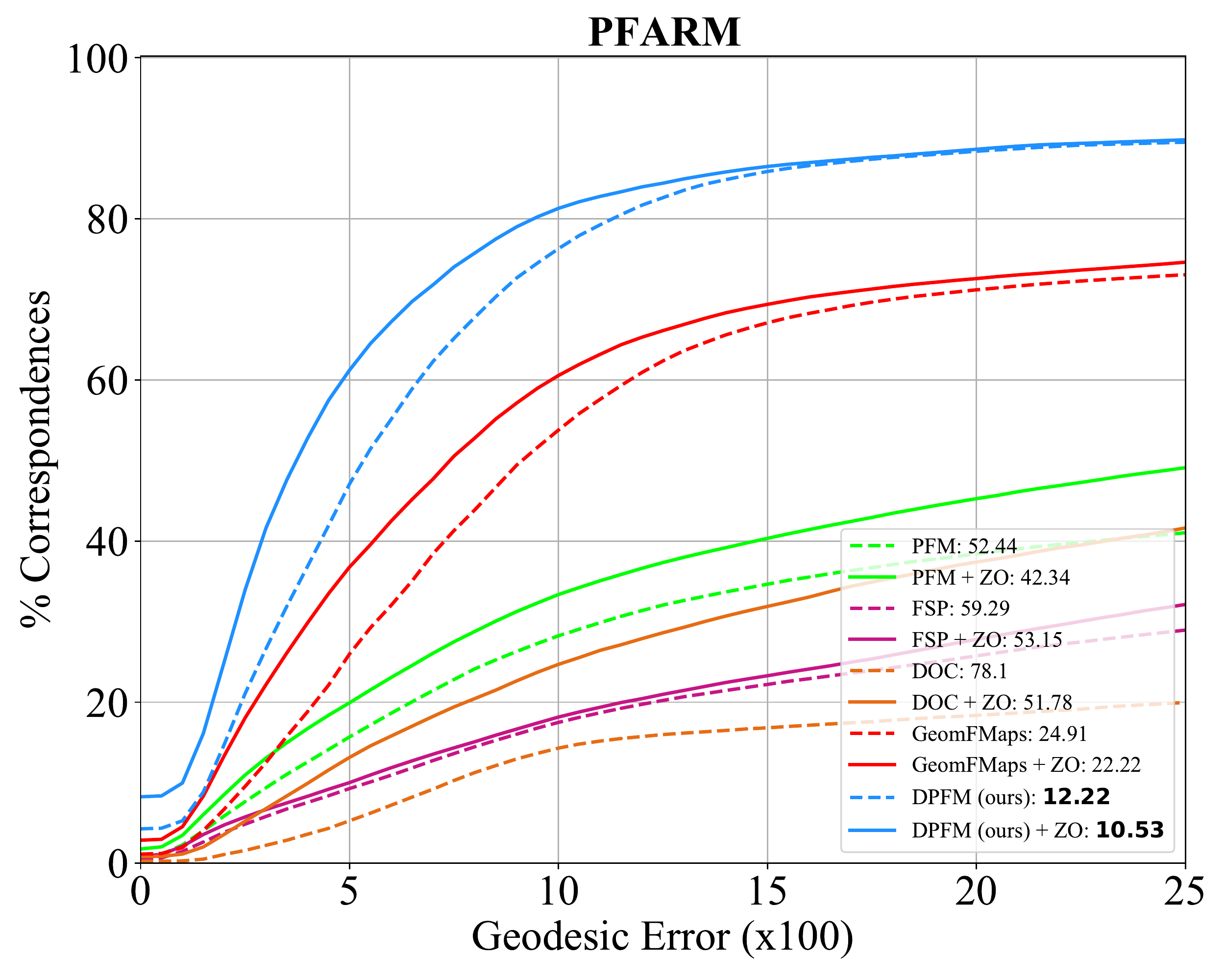}
\end{center}
\vspace{-2mm}
   \caption{Results on the PFARM dataset. Our method is  robust to significant changes in mesh discretization and outperforms all the baselines, whether learning-based or axiomatic. \vspace{-1mm}}
\label{fig:princeton_farm}
\end{figure}

Figure \ref{fig:princeton_farm} compares the performance of all methods on the \textbf{PFARM} dataset.  As the size of this dataset is small, and is not adapted to training, we only used the pretrained networks on CUTS for both our method and GeomFMaps.

\begin{figure}[]
\vspace{-6mm}
\begin{center}
\includegraphics[width=0.95\linewidth]{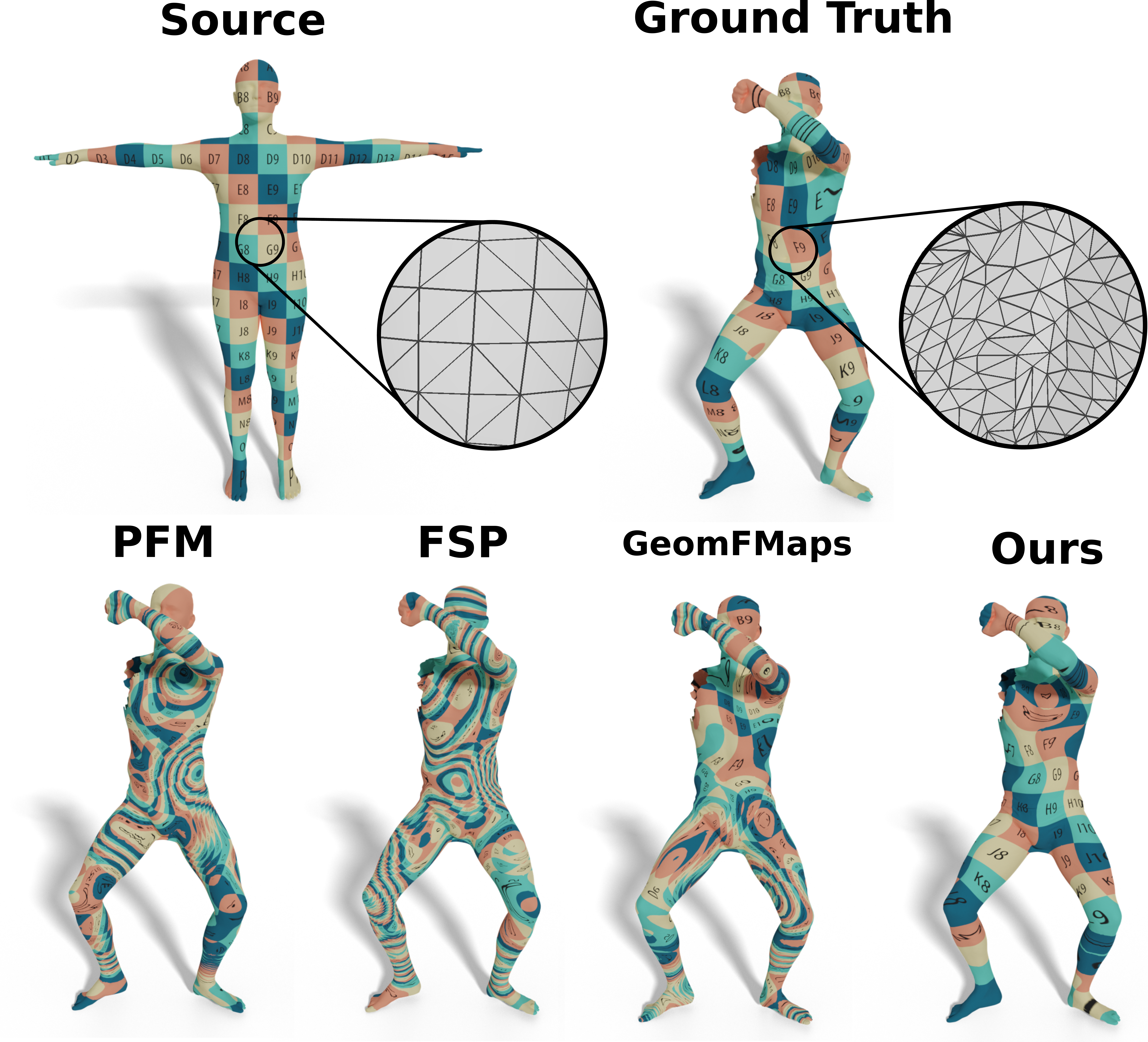}
\end{center}
\vspace{-6mm}
\caption{Qualitative results on \textbf{PFARM} using texture transfer. Only our method produces visually plausible results. Additional visualizations for \textbf{SHREC'16} and \textbf{CP2P} are provided in the supplementary.\vspace{-6mm}}
\label{fig:quali_farm}
\end{figure}

We observe that competing methods that rely on handcrafted features like SHOT, tend to \emph{overfit} to the mesh connectivity and as a result see a significant performance drop on meshes with variable sampling and connectivity like \textbf{PFARM}.  In contrast, our network which is \emph{pre-trained on} \textbf{CUTS} from \textbf{SHREC'16} generalises significantly better and has considerably lower errors than the competitors. Figure \ref{fig:quali_farm} shows a visualization of the predicted maps from all methods. 
It can be seen that only our method yields visually plausible correspondences on the shapes from \textbf{PFARM}.

\subsection{Part To Part Shape Matching}
\label{subsec:p2p_exp}
\begin{figure}[]
\begin{center}
\includegraphics[width=0.92\linewidth]{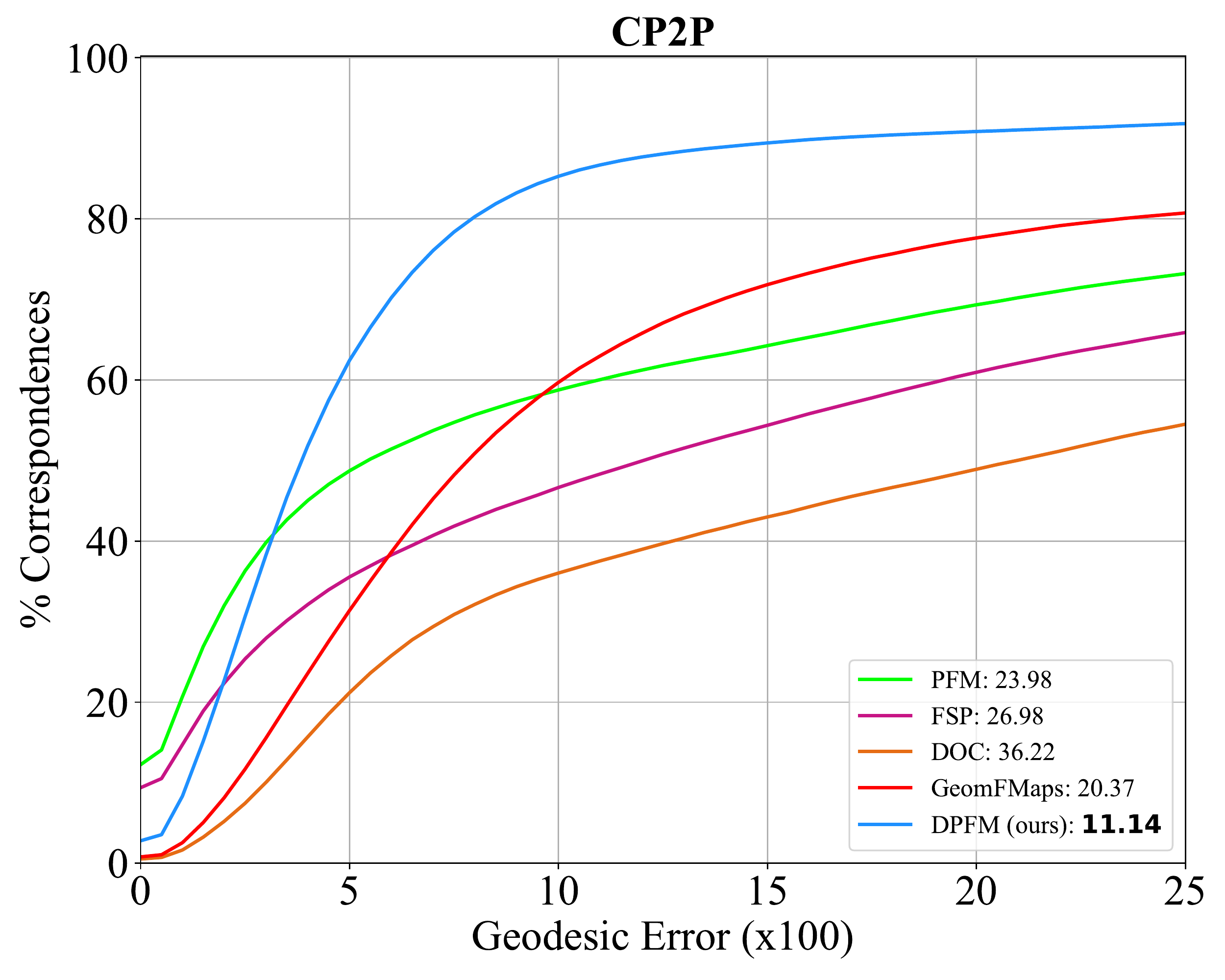}
\end{center}
\vspace{-3mm}
   \caption{Quantitative results of the different methods on the CP2P dataset. Our method outperforms all baselines. \vspace{-3mm}} 
\label{fig:princeton_p2p}
\end{figure}

We evaluate our method on the challenging task of partial-to-partial shape matching, where only a portion of the source shape is mapped to a portion of the target shape. 
We compare our method against PFM \cite{Rodol2016}, the formulation of FSP adapted to the partial-to-partial setting \cite{Litany2017}, DOC \cite{Cosmo2016}, and the learning-based baseline GeomFMaps \cite{Donati2020} on the \textbf{CP2P} dataset. We apply PFM in both directions (source to target and target to source), thereby generating a predicted region of overlap on both source and target shapes. Since DOC \cite{Cosmo2016} already optimizes for two overlap regions, we use their method as-is by initializing with SHOT descriptors. In order to evaluate the quality of matching, we show the error plot for correspondences \emph{only in the overlapping region}. Figure \ref{fig:princeton_p2p} shows the superiority of our method over the baselines. We also plot the IOU of the predicted overlap region to assess the quality of the overlap region detection, our method gives superior results (see the supplementary). 

\section{Conclusion, Limitations, and Future work}
\label{sec:conclusion}



In this paper, we present a learning framework specifically aimed for partial non-rigid shape matching. We exploit the versatility and accuracy of powerful geometric feature extractors along with the strong regularization provided by the functional map formulation. Key to our design is an attention-based feature refiner, which ensures cross-talk between features on the two shapes and is therefore crucial in the context of partial shape correspondence. Our method gives a significant improvement upon existing baselines, especially highlighted in challenging scenarios of considerable variability in mesh discretization.

Our approach still has some limitations and leads to possible exciting future work. Firstly, similar to many existing methods in practice, we assume that the shapes undergo a near-isometric deformation on the overlap region, a restrictive assumption that could be potentially lifted. In addition, extending our framework to more challenging representations such as noisy point clouds and graphs is also a fascinating area for future research.

\vspace{2mm}
\textbf{Acknowledgements}
We gratefully acknowledge anonymous reviewers for their valuable suggestions. We also thank Rui Xiang for sharing code and help in executing the baselines as well as Simone Melzi and Emanuele Rodol\`a for fruitful discussions. Parts of this work were supported by the ERC Starting Grants No. 758800 (EXPROTEA) and the ANR AI Chair AIGRETTE.

{\small
\bibliographystyle{ieee_fullname}
\bibliography{egbib}

\begin{thebibliography}{10}\itemsep=-1pt

\bibitem{Arbel2019}
Nadav~Yehonatan Arbel, Ayellet Tal, and Lihi Zelnik-Manor.
\newblock Partial correspondence of 3d shapes using properties of the
  nearest-neighbor field.
\newblock {\em Computers {\&} Graphics}, 82:183--192, Aug. 2019.

\bibitem{bellekens2014survey}
Ben Bellekens, Vincent Spruyt, Rafael Berkvens, and Maarten Weyn.
\newblock A survey of rigid 3d pointcloud registration algorithms.
\newblock In {\em AMBIENT 2014: the Fourth International Conference on Ambient
  Computing, Applications, Services and Technologies, August 24-28, 2014, Rome,
  Italy}, pages 8--13, 2014.

\bibitem{biasotti2016recent}
Silvia Biasotti, Andrea Cerri, Alex Bronstein, and Michael Bronstein.
\newblock Recent trends, applications, and perspectives in 3d shape similarity
  assessment.
\newblock {\em Computer Graphics Forum}, 35(6):87--119, 2016.

\bibitem{boscaini2016}
Davide Boscaini, Jonathan Masci, Emanuele Rodol\`{a}, and Michael Bronstein.
\newblock Learning shape correspondence with anisotropic convolutional neural
  networks.
\newblock In D. Lee, M. Sugiyama, U. Luxburg, I. Guyon, and R. Garnett,
  editors, {\em Advances in Neural Information Processing Systems}, volume~29.
  Curran Associates, Inc., 2016.

\bibitem{bouaziz2013sparse}
Sofien Bouaziz, Andrea Tagliasacchi, and Mark Pauly.
\newblock Sparse iterative closest point.
\newblock In {\em Computer graphics forum}, volume~32, pages 113--123. Wiley
  Online Library, 2013.

\bibitem{bronstein2017geometric}
Michael~M Bronstein, Joan Bruna, Yann LeCun, Arthur Szlam, and Pierre
  Vandergheynst.
\newblock Geometric deep learning: going beyond euclidean data.
\newblock {\em IEEE Signal Processing Magazine}, 34(4):18--42, 2017.

\bibitem{cao2020comprehensive}
Wenming Cao, Zhiyue Yan, Zhiquan He, and Zhihai He.
\newblock A comprehensive survey on geometric deep learning.
\newblock {\em IEEE Access}, 8:35929--35949, 2020.

\bibitem{castellani20203d}
Umberto Castellani and Adrien Bartoli.
\newblock 3d shape registration.
\newblock In {\em 3D Imaging, Analysis and Applications}, pages 353--411.
  Springer, 2020.

\bibitem{chetverikov2002trimmed}
Dmitry Chetverikov, Dmitry Svirko, Dmitry Stepanov, and Pavel Krsek.
\newblock The trimmed iterative closest point algorithm.
\newblock In {\em Object recognition supported by user interaction for service
  robots}, volume~3, pages 545--548. IEEE, 2002.

\bibitem{choy2019fully}
Christopher Choy, Jaesik Park, and Vladlen Koltun.
\newblock Fully convolutional geometric features.
\newblock In {\em Proceedings of the IEEE/CVF International Conference on
  Computer Vision}, pages 8958--8966, 2019.

\bibitem{cosmo2016shrec}
Luca Cosmo, Emanuele Rodola, Michael~M Bronstein, Andrea Torsello, Daniel
  Cremers, and Y Sahillioglu.
\newblock Shrec’16: Partial matching of deformable shapes.
\newblock {\em Proc. 3DOR}, 2(9):12, 2016.

\bibitem{Cosmo2016}
Luca Cosmo, Emanuele Rodola, Jonathan Masci, Andrea Torsello, and Michael~M.
  Bronstein.
\newblock Matching deformable objects in clutter.
\newblock In {\em 2016 Fourth International Conference on 3D Vision (3DV)}.
  {IEEE}, Oct. 2016.

\bibitem{deng2018ppfnet}
Haowen Deng, Tolga Birdal, and Slobodan Ilic.
\newblock {PPF}net: Global context aware local features for robust 3d point
  matching.
\newblock In {\em Proceedings of the IEEE conference on computer vision and
  pattern recognition}, pages 195--205, 2018.

\bibitem{Donati2020}
Nicolas Donati, Abhishek Sharma, and Maks Ovsjanikov.
\newblock Deep geometric functional maps: Robust feature learning for shape
  correspondence.
\newblock In {\em 2020 {IEEE}/{CVF} Conference on Computer Vision and Pattern
  Recognition ({CVPR})}. {IEEE}, June 2020.

\bibitem{du2011robust}
Shaoyi Du, Jihua Zhu, Nanning Zheng, Yuehu Liu, and Ce Li.
\newblock Robust iterative closest point algorithm for registration of point
  sets with outliers.
\newblock {\em Optical Engineering}, 50(8):087001, 2011.

\bibitem{gojcic2019perfect}
Zan Gojcic, Caifa Zhou, Jan~D Wegner, and Andreas Wieser.
\newblock The perfect match: 3d point cloud matching with smoothed densities.
\newblock In {\em Proceedings of the IEEE/CVF Conference on Computer Vision and
  Pattern Recognition}, pages 5545--5554, 2019.

\bibitem{graham20183d}
Benjamin Graham, Martin Engelcke, and Laurens Van Der~Maaten.
\newblock 3d semantic segmentation with submanifold sparse convolutional
  networks.
\newblock In {\em Proceedings of the IEEE conference on computer vision and
  pattern recognition}, pages 9224--9232, 2018.

\bibitem{halimi2019unsupervised}
Oshri Halimi, Or Litany, Emanuele Rodola, Alex~M Bronstein, and Ron Kimmel.
\newblock Unsupervised learning of dense shape correspondence.
\newblock In {\em Proceedings of the IEEE/CVF Conference on Computer Vision and
  Pattern Recognition}, pages 4370--4379, 2019.

\bibitem{huang2021predator}
Shengyu Huang, Zan Gojcic, Mikhail Usvyatsov, Andreas Wieser, and Konrad
  Schindler.
\newblock Predator: Registration of 3d point clouds with low overlap.
\newblock In {\em Proceedings of the IEEE/CVF Conference on Computer Vision and
  Pattern Recognition}, pages 4267--4276, 2021.

\bibitem{Kim2011}
Vladimir~G. Kim, Yaron Lipman, and Thomas Funkhouser.
\newblock Blended intrinsic maps.
\newblock {\em {ACM} Transactions on Graphics}, 30(4):1--12, July 2011.

\bibitem{kingma2017adam}
Diederik~P. Kingma and Jimmy Ba.
\newblock Adam: A method for stochastic optimization, 2017.

\bibitem{kipf2017semi}
Thomas~N. Kipf and Max Welling.
\newblock Semi-supervised classification with graph convolutional networks.
\newblock In {\em International Conference on Learning Representations (ICLR)},
  2017.

\bibitem{Kirgo2020}
Maxime Kirgo, Simone Melzi, Giuseppe Patan{\`{e}}, Emanuele Rodol{\`{a}}, and
  Maks Ovsjanikov.
\newblock Wavelet-based heat kernel derivatives: Towards informative localized
  shape analysis.
\newblock {\em Computer Graphics Forum}, 40(1):165--179, Nov. 2020.

\bibitem{kovnatsky2013coupled}
Artiom Kovnatsky, Michael~M Bronstein, Alexander~M Bronstein, Klaus Glashoff,
  and Ron Kimmel.
\newblock Coupled quasi-harmonic bases.
\newblock In {\em Computer Graphics Forum}, volume~32, pages 439--448. Wiley
  Online Library, 2013.

\bibitem{Li_2020_CVPR}
Qinsong Li, Shengjun Liu, Ling Hu, and Xinru Liu.
\newblock Shape correspondence using anisotropic chebyshev spectral cnns.
\newblock In {\em Proceedings of the IEEE/CVF Conference on Computer Vision and
  Pattern Recognition (CVPR)}, June 2020.

\bibitem{litany2017deep}
Or Litany, Tal Remez, Emanuele Rodola, Alex Bronstein, and Michael Bronstein.
\newblock Deep functional maps: Structured prediction for dense shape
  correspondence.
\newblock In {\em Proceedings of the IEEE international conference on computer
  vision}, pages 5659--5667, 2017.

\bibitem{Litany2017}
O. Litany, E. Rodol{\`{a}}, A.~M. Bronstein, and M.~M. Bronstein.
\newblock Fully spectral partial shape matching.
\newblock {\em Computer Graphics Forum}, 36(2):247--258, May 2017.

\bibitem{Litany2016}
O. Litany, E. Rodol{\`{a}}, A.~M. Bronstein, M.~M. Bronstein, and D. Cremers.
\newblock Non-rigid puzzles.
\newblock {\em Computer Graphics Forum}, 35(5):135--143, Aug. 2016.

\bibitem{SMPL_2015}
Matthew Loper, Naureen Mahmood, Javier Romero, Gerard Pons-Moll, and Michael~J.
  Black.
\newblock {SMPL}: A skinned multi-person linear model.
\newblock {\em ACM Trans. Graphics (Proc. SIGGRAPH Asia)}, 34(6):248:1--248:16,
  Oct. 2015.

\bibitem{marin2020farm}
Riccardo Marin, Simone Melzi, Emanuele Rodola, and Umberto Castellani.
\newblock Farm: Functional automatic registration method for 3d human bodies.
\newblock In {\em Computer Graphics Forum}, volume~39, pages 160--173. Wiley
  Online Library, 2020.

\bibitem{masci2015geodesic}
Jonathan Masci, Davide Boscaini, Michael Bronstein, and Pierre Vandergheynst.
\newblock Geodesic convolutional neural networks on riemannian manifolds.
\newblock In {\em Proceedings of the IEEE international conference on computer
  vision workshops}, pages 37--45, 2015.

\bibitem{shrec19connectivity}
S. Melzi, R. Marin, E. Rodolà, U. Castellani, J. Ren, A. Poulenard, P. Wonka,
  and M. Ovsjanikov.
\newblock {Matching Humans with Different Connectivity}.
\newblock In Silvia Biasotti, Guillaume Lavoué, and Remco Veltkamp, editors,
  {\em Eurographics Workshop on 3D Object Retrieval}. The Eurographics
  Association, 2019.

\bibitem{Melzi2019}
Simone Melzi, Jing Ren, Emanuele Rodol{\`{a}}, Abhishek Sharma, Peter Wonka,
  and Maks Ovsjanikov.
\newblock {ZoomOut: Spectral Upsampling for Efficient Shape Correspondence}.
\newblock {\em {ACM} Transactions on Graphics}, 38(6):1--14, Nov. 2019.

\bibitem{monti2017geometric}
Federico Monti, Davide Boscaini, Jonathan Masci, Emanuele Rodola, Jan Svoboda,
  and Michael~M Bronstein.
\newblock Geometric deep learning on graphs and manifolds using mixture model
  cnns.
\newblock In {\em Proceedings of the IEEE conference on computer vision and
  pattern recognition}, pages 5115--5124, 2017.

\bibitem{nwankpa2018activation}
Chigozie Nwankpa, Winifred Ijomah, Anthony Gachagan, and Stephen Marshall.
\newblock Activation functions: Comparison of trends in practice and research
  for deep learning.
\newblock {\em arXiv preprint arXiv:1811.03378}, 2018.

\bibitem{ovsjanikov2012functional}
Maks Ovsjanikov, Mirela Ben-Chen, Justin Solomon, Adrian Butscher, and Leonidas
  Guibas.
\newblock Functional maps: a flexible representation of maps between shapes.
\newblock {\em ACM Transactions on Graphics (TOG)}, 31(4):1--11, 2012.

\bibitem{pais20203dregnet}
G~Dias Pais, Srikumar Ramalingam, Venu~Madhav Govindu, Jacinto~C Nascimento,
  Rama Chellappa, and Pedro Miraldo.
\newblock 3dregnet: A deep neural network for 3d point registration.
\newblock In {\em Proceedings of the IEEE/CVF conference on computer vision and
  pattern recognition}, pages 7193--7203, 2020.

\bibitem{NEURIPS2019_9015}
Adam Paszke, Sam Gross, Francisco Massa, Adam Lerer, James Bradbury, Gregory
  Chanan, Trevor Killeen, Zeming Lin, Natalia Gimelshein, Luca Antiga, Alban
  Desmaison, Andreas Kopf, Edward Yang, Zachary DeVito, Martin Raison, Alykhan
  Tejani, Sasank Chilamkurthy, Benoit Steiner, Lu Fang, Junjie Bai, and Soumith
  Chintala.
\newblock Pytorch: An imperative style, high-performance deep learning library.
\newblock In H. Wallach, H. Larochelle, A. Beygelzimer, F. d\textquotesingle
  Alch\'{e}-Buc, E. Fox, and R. Garnett, editors, {\em Advances in Neural
  Information Processing Systems 32}, pages 8024--8035. Curran Associates,
  Inc., 2019.

\bibitem{pishchulin2017building}
Leonid Pishchulin, Stefanie Wuhrer, Thomas Helten, Christian Theobalt, and
  Bernt Schiele.
\newblock Building statistical shape spaces for 3d human modeling.
\newblock {\em Pattern Recognition}, 67:276--286, 2017.

\bibitem{poulenard2018multi}
Adrien Poulenard and Maks Ovsjanikov.
\newblock Multi-directional geodesic neural networks via equivariant
  convolution.
\newblock {\em ACM Transactions on Graphics (TOG)}, 37(6):1--14, 2018.

\bibitem{pointnet++}
Charles~R. Qi, Li Yi, Hao Su, and Leonidas~J. Guibas.
\newblock Pointnet++: Deep hierarchical feature learning on point sets in a
  metric space.
\newblock In {\em Proceedings of the 31st International Conference on Neural
  Information Processing Systems}, NIPS'17, page 5105–5114, Red Hook, NY,
  USA, 2017. Curran Associates Inc.

\bibitem{rampini2019correspondence}
Arianna Rampini, Irene Tallini, Maks Ovsjanikov, Alex~M Bronstein, and Emanuele
  Rodola.
\newblock Correspondence-free region localization for partial shape similarity
  via hamiltonian spectrum alignment.
\newblock In {\em 2019 International Conference on 3D Vision (3DV)}, pages
  37--46. IEEE, 2019.

\bibitem{Ren2019}
Jing Ren, Mikhail Panine, Peter Wonka, and Maks Ovsjanikov.
\newblock Structured regularization of functional map computations.
\newblock {\em Computer Graphics Forum}, 38(5):39--53, Aug. 2019.

\bibitem{rodola2012game}
Emanuele Rodola, Alex~M Bronstein, Andrea Albarelli, Filippo Bergamasco, and
  Andrea Torsello.
\newblock A game-theoretic approach to deformable shape matching.
\newblock In {\em 2012 IEEE Conference on Computer Vision and Pattern
  Recognition}, pages 182--189. IEEE, 2012.

\bibitem{Rodol2016}
E. Rodol{\`{a}}, L. Cosmo, M.~M. Bronstein, A. Torsello, and D. Cremers.
\newblock Partial functional correspondence.
\newblock {\em Computer Graphics Forum}, 36(1):222--236, Feb. 2016.

\bibitem{RFB15a}
O. Ronneberger, P.Fischer, and T. Brox.
\newblock U-net: Convolutional networks for biomedical image segmentation.
\newblock In {\em Medical Image Computing and Computer-Assisted Intervention
  (MICCAI)}, volume 9351 of {\em LNCS}, pages 234--241. Springer, 2015.
\newblock (available on arXiv:1505.04597 [cs.CV]).

\bibitem{roufosse2019unsupervised}
Jean-Michel Roufosse, Abhishek Sharma, and Maks Ovsjanikov.
\newblock Unsupervised deep learning for structured shape matching.
\newblock In {\em Proceedings of the IEEE/CVF International Conference on
  Computer Vision}, pages 1617--1627, 2019.

\bibitem{rusinkiewicz2001efficient}
Szymon Rusinkiewicz and Marc Levoy.
\newblock Efficient variants of the icp algorithm.
\newblock In {\em Proceedings third international conference on 3-D digital
  imaging and modeling}, pages 145--152. IEEE, 2001.

\bibitem{sahilliouglu2020recent}
Yusuf Sahillio{\u{g}}lu.
\newblock Recent advances in shape correspondence.
\newblock {\em The Visual Computer}, 36(8):1705--1721, 2020.

\bibitem{salti2014shot}
Samuele Salti, Federico Tombari, and Luigi Di~Stefano.
\newblock Shot: Unique signatures of histograms for surface and texture
  description.
\newblock {\em Computer Vision and Image Understanding}, 125:251--264, 2014.

\bibitem{SharmaO20}
Abhishek Sharma and Maks Ovsjanikov.
\newblock Weakly supervised deep functional maps for shape matching.
\newblock {\em Advances in Neural Information Processing Systems}, 33, 2020.

\bibitem{sharp2021diffusionnet}
Nicholas Sharp, Souhaib Attaiki, Keenan Crane, and Maks Ovsjanikov.
\newblock Diffusionnet: Discretization agnostic learning on surfaces.
\newblock {\em arXiv preprint arXiv:2012.00888}, 2021.

\bibitem{sumner2004deformation}
Robert~W Sumner and Jovan Popovi{\'c}.
\newblock Deformation transfer for triangle meshes.
\newblock {\em ACM Transactions on graphics (TOG)}, 23(3):399--405, 2004.

\bibitem{thomas2019kpconv}
Hugues Thomas, Charles~R Qi, Jean-Emmanuel Deschaud, Beatriz Marcotegui,
  Fran{\c{c}}ois Goulette, and Leonidas~J Guibas.
\newblock Kpconv: Flexible and deformable convolution for point clouds.
\newblock In {\em Proceedings of the IEEE/CVF International Conference on
  Computer Vision}, pages 6411--6420, 2019.

\bibitem{ulyanov2016instance}
Dmitry Ulyanov, Andrea Vedaldi, and Victor Lempitsky.
\newblock Instance normalization: The missing ingredient for fast stylization.
\newblock {\em arXiv preprint arXiv:1607.08022}, 2016.

\bibitem{van2011survey}
Oliver Van~Kaick, Hao Zhang, Ghassan Hamarneh, and Daniel Cohen-Or.
\newblock A survey on shape correspondence.
\newblock {\em Computer Graphics Forum}, 30(6):1681--1707, 2011.

\bibitem{attentionneed}
Ashish Vaswani, Noam Shazeer, Niki Parmar, Jakob Uszkoreit, Llion Jones,
  Aidan~N Gomez, {\L}ukasz Kaiser, and Illia Polosukhin.
\newblock Attention is all you need.
\newblock In {\em Advances in neural information processing systems}, pages
  5998--6008, 2017.

\bibitem{attention2018graph}
Petar Veli{\v{c}}kovi{\'c}, Guillem Cucurull, Arantxa Casanova, Adriana Romero,
  Pietro Lio, and Yoshua Bengio.
\newblock Graph attention networks.
\newblock {\em arXiv preprint arXiv:1710.10903}, 2017.

\bibitem{wang2019prnet}
Yue Wang and Justin~M Solomon.
\newblock {PR}net: Self-supervised learning for partial-to-partial
  registration.
\newblock {\em arXiv preprint arXiv:1910.12240}, 2019.

\bibitem{wiersma2020cnns}
Ruben Wiersma, Elmar Eisemann, and Klaus Hildebrandt.
\newblock Cnns on surfaces using rotation-equivariant features.
\newblock {\em ACM Transactions on Graphics (TOG)}, 39(4):92--1, 2020.

\bibitem{MLP}
E. Wilson and D.W. Tufts.
\newblock Multilayer perceptron design algorithm.
\newblock In {\em Proceedings of IEEE Workshop on Neural Networks for Signal
  Processing}, pages 61--68, 1994.

\bibitem{Wu2020}
Yan Wu, Jun Yang, and Jinlong Zhao.
\newblock Partial 3d shape functional correspondence via fully spectral
  eigenvalue alignment and upsampling refinement.
\newblock {\em Computers {\&} Graphics}, 92:99--113, Nov. 2020.

\bibitem{Xiang_2021_CVPR}
Rui Xiang, Rongjie Lai, and Hongkai Zhao.
\newblock A dual iterative refinement method for non-rigid shape matching.
\newblock In {\em Proceedings of the IEEE/CVF Conference on Computer Vision and
  Pattern Recognition (CVPR)}, pages 15930--15939, June 2021.

\bibitem{Xie_2020}
Saining Xie, Jiatao Gu, Demi Guo, Charles~R. Qi, Leonidas Guibas, and Or
  Litany.
\newblock Pointcontrast: Unsupervised pre-training for 3d point cloud
  understanding.
\newblock {\em Lecture Notes in Computer Science}, page 574–591, 2020.

\bibitem{xu2015empirical}
Bing Xu, Naiyan Wang, Tianqi Chen, and Mu Li.
\newblock Empirical evaluation of rectified activations in convolutional
  network.
\newblock {\em arXiv preprint arXiv:1505.00853}, 2015.

\bibitem{zeng20173dmatch}
Andy Zeng, Shuran Song, Matthias Nie{\ss}ner, Matthew Fisher, Jianxiong Xiao,
  and Thomas Funkhouser.
\newblock 3dmatch: Learning local geometric descriptors from rgb-d
  reconstructions.
\newblock In {\em Proceedings of the IEEE conference on computer vision and
  pattern recognition}, pages 1802--1811, 2017.

\end{thebibliography}
}


\clearpage








\onecolumn


\begin{center}
  {\Large \bf DPFM: Deep Partial Functional Maps -  Supplementary Material \par}
   \vspace*{24pt}
   {
      \large
      \lineskip .5em
      \begin{tabular}[t]{c}
          Souhaib Attaiki \qquad Gautam Pai  \qquad Maks Ovsjanikov\\
LIX, École Polytechnique, IP Paris\\
         \vspace*{1pt}\\
      \end{tabular}
      \par
      }
      \vskip .5em
      \vspace*{12pt}
\end{center}

\begin{multicols}{2}

\setcounter{equation}{0}
\setcounter{figure}{0}
\setcounter{table}{0}
\setcounter{page}{1}

\appendix

In this supplementary materials, we collect all the results and discussions, which, due to the page limit, could not find space in the main manuscript.

Specifically, we provide a discussion about functional maps in the partial setting in Section \ref{sec:partial_theory}. Next, we present our unsupervised formulation and show one additional result in this setting in Section \ref{sec:usnup}. We provide implementations details for our experiments in Section \ref{sec:imp_details}. In Section \ref{sec:data_viz}, additional information about the used datasets and some example shapes are visualized. Section \ref{sec:refinement_module} provides additional insights and quantitative evaluation of our Cross-Attention Refinement module. Additional quantitative and qualitative evaluations of our method are provided in Section \ref{sec:quan_eva} and \ref{sec:qual_eva} respectively. Finally, an ablation study on the component of our architecture is provided in Section \ref{sec:abla}.

\section{Discussion about functional maps in the partial setting}
\label{sec:partial_theory}

In this section, we provide some analysis of the partial non-rigid matching problem, especially
within the functional maps framework. Our main goal is to establish the necessary conditions under
which the feature extraction network on the two shapes must be ``aware'' of the other
shape. Specifically, we show that this communication across feature extraction networks on the two
shapes must be necessary for partial-to-partial matching, within the functional maps representation.

\subsection{Partial to full matching}
We start with the simpler case of partial to full matching. Namely, suppose a partial shape $\X$ is
being matched to the full shape $\Y$. In this case, there exists a point-to-point map
$T: \X \rightarrow \Y$ so that for each point on $\X$ there is a corresponding point on $\Y$. In the
discrete setting, this map can be written as a binary matrix $\Pi_{\X\Y}$ where
$\Pi_{\X\Y}(i,j) = 1$ if and only if $T(i) = j$. Note that $\Pi_{\X\Y}$ has exactly one value 1 per
row. The corresponding functional map $C_{\Y\X}$ maps functions on $\Y$ to functions on $\X$ and, in
the reduced Laplacian basis can be written as: $C_{\Y\X} = \Phi^{+}_\X \Pi_{\X\Y} \Phi_{\Y}.$

Now suppose that a network $\F^{*}$ is a \emph{\textbf{perfect feature extractor}} in the following
sense: given a shape, $\X$, it associates to each point on $\X$ a unique non-zero descriptor vector
that is moreover invariant under different possible transformations (including shape
deformations, or part removal) of the shape $\X$. We let $\F^{*}(\X) = D_{\X}$ where the
$i^{\text{th}}$ row of $D_{\X}$ corresponds to the descriptor of vertex $i$ on $\X$.

If $\F^{*}$ is a perfect feature extractor and $\Pi_{\X\Y}$ is the underlying ground truth map
between $\X$ and $\Y$, then by definition, we have $D_{\X} = \Pi_{\X\Y} D_{\Y}$. Remark that in
order for this equation to hold, the feature extractor simply needs to be invariant under the shape
partiality and does not need to be dependent on the map $\Pi_{\X\Y}$. For example, $D_{\X}$ could
store, for every point, the index of the corresponding point on some template shape.

If $D_{\X} = \Pi_{\X\Y} D_{\Y}$ then in the reduced basis we have $\Phi_{\X}^{+} D_{\X} =
\Phi_{\X}^{+} \Pi_{\X\Y} D_{\Y}$. Moreover, using the standard assumption in the functional maps
framework, that the descriptor matrix $D_{\Y}$ lies within the span of the reduced Laplacian basis
on $\Y$ we have $D_{\Y} = \Phi_{\Y} \Phi_{\Y}^{+} D_{\Y}$. This implies:
\begin{align*}
  \Phi_{\X}^{+} D_{\X} &=
                         \Phi_{\X}^{+} \Pi_{\X\Y} D_{\Y}  \\
                       &=  \Phi_{\X}^{+} \Pi_{\X\Y}  \Phi_{\Y} \Phi_{\Y}^{+} D_{\Y} \\
  &= \mathbf{C}_{\Y\X} \mathbf{D}_{\Y},
\end{align*}
where by definition $\mathbf{C}_{\Y\X} = \Phi_{\X}^{+} \Pi_{\X\Y}  \Phi_{\Y}$ and $\mathbf{D}_{\Y} =
\Phi_{\Y}^{+} D_{\Y}$.

We therefore conclude that if
$\mathbf{D}_{\X} = \Phi_{\X}^{+} D_{\X}$, then under the above
assumptions (i.e., having a perfect feature extractor and descriptors
within the span of the basis), we have
$\| \mathbf{C}_{\Y\X} \mathbf{D}_{\Y} - \mathbf{D}_{\X}\| = 0,$ and if
the linear system is invertible the functional map $\mathbf{C}_{\Y\X}$
can be recovered via
$\mathbf{C}_{\Y\X} = \argmin_{\mathbf{X}} \|\mathbf{X} \mathbf{D}_{\Y}
- \mathbf{D}_{\X}\|^2_F.$

It then follows ``communication'' between feature extraction on the two shapes is not strictly
required to recover the underlying functional map in the case of partial to full matching.

\noindent \textbf{Note:} Observe that the argument above did not make assumptions on the rank of the functional
map matrix or on the number of basis functions. Consider some part that exists on the full shape
$\Y$ and \emph{does not exist} on the partial shape $\X$. If $f$ is the descriptor that associates a
feature value only to points on that part, the equation $C_{\Y\X} \Phi^{+}_{\Y} f_{\Y}
=\Phi^{+}_{\X}  f_{\X} $ can hold even if $f_{\X} = 0$ but $f_{\Y} \ne 0$. I.e., $\mathbf{C}
\mathbf{a} = \mathbf{b}$ and $\mathbf{b}=0$ does not imply that $\mathbf{a} = 0$.

\subsection{Full-to-partial and partial-to-partial matching}
Consider now full-to-partial or partial-to-partial matching. Here, unlike the case above, given a
source shape $\X$ and a target shape $\Y$, there exists a mapping $T: S \subset \X \rightarrow \Y$
\emph{only for a subset of  points} $S \subset \X$.

We can still represent the mapping $T$ as a binary matrix $\Pi_{\X\Y}$, s.t., $\Pi_{\X\Y}(i,j) = 1$
if and only if $T(i) = j$. However, in this case, the matrix $\Pi_{\X\Y}$ will have rows that are
entirely zero, for points outside of the subset $S$ that don't have a map onto $\Y$.

Observe that in this case, if $D_{\X}, D_{\Y}$ are features obtained by a perfect feature extractor
$\mathcal{F}^{*}$ as defined above, then we cannot have $D_{\X} = \Pi_{\X\Y} D_{\Y}$. This is
because the matrix $\Pi_{\X\Y}$ will map features of points outside of $S$ onto the zero
vector. I.e., for any point $i \notin S$ the corresponding row of the matrix $\Pi_{\X\Y} D_{\Y}$
will be exactly zero, whereas $D_{\X}$ by assumption is not a zero vector.

While the ``standard'' equation $D_{\X} = \Pi_{\X\Y} D_{\Y}$ does not hold, a modified version can
easily be seen to hold. Let $P_{\X}$ by the binary matrix that is identity on $S$ and zeros
elsewhere. I.e., $P(i,i) = 1$ if and only if $i\in S$. Then, again under the assumptions of a
perfect feature extractor we have:
\begin{align*}
  P_{\X} D_{\X} = \Pi_{\X\Y} D_{\Y}.
\end{align*}

If the descriptors are within the span of the Laplacian basis, this implies:
\begin{align*}
  \Phi^{+}_{\X}  P_{\X} D_{\X} &= \Phi^{+}_{\X} \Pi_{\X\Y} \Phi^{+}_{\Y} \Phi^{+}_{\Y} D_{\Y} \\
  &= \mathbf{C}_{\Y\X} \mathbf{D}_{\Y}.
\end{align*}

I.e., we have $\| \mathbf{C}_{\Y\X} \mathbf{D}_{\Y} - \mathbf{D}_{\X}\| = 0$. However, crucially, in
this case $\mathbf{D}_{\X} = \Phi^{+}_{\X} P_{\X} D_{\X}$, where $P_{\X}$ is the projection matrix
onto the set $S$. We stress that the set $S$ \emph{\textbf{depends on the underlying map}} (i.e.,
the target shape) and therefore, unless the feature extractor is aware of the target shape being
mapped to, it cannot extract features for which
$\| \mathbf{C}_{\Y\X} \mathbf{D}_{\Y} - \mathbf{D}_{\X}\| = 0$.

We, therefore, conclude that ``communication'' between feature extraction on the two shapes is 
required to recover the underlying functional map from the feature equation in the case of full to
partial or partial to partial matching.

\noindent \textbf{Note:} Consider some part that exists on the shape $\X$ and \emph{does not exist}
on shape $\Y$. If $f$ is the descriptor that associates a feature value only to points on that part,
the equation $C_{\Y\X} \Phi^{+}_{\Y} f_{\Y} =\Phi^{+}_{\X} f_{\X} $ where $f_{\Y} = 0$ can hold only
if $f_{\X} = 0$. But this is only possible if the feature extractor on $\X$ has access to the
information about features on $\Y$ for otherwise it would extract non-zero features $f_{\X}$ for
that part. This confirms the above interpretation that in this case ``communication'' between
feature extraction on the two shapes is necessary.

\section{Unsupervised partial shape matching}
\label{sec:usnup}
In our main document, we presented losses to train the network in the supervised setting. Here, we present a loss that can be used in the unsupervised setting for partial-to-full matching, and that works by promoting structural properties of the functional map. It should be noted that in the unsupervised case, we predict the functional map in both directions, i.e partial-to-full and full-to-partial, by applying our Regularized FMap module in the following manner (we follow the same notation as the main text, the full shape will be denoted by shape 1, and the partial shape will be denoted shape 2):
$$C_{12} = \argmin_{C} \|C \mathbf{A} - \mathbf{B}\|_F^2 + \lambda \sum_{ij} C_{ij}^2 M_{ij}^{12}$$
$$C_{21} = \argmin_{C} \|C \mathbf{B} - \mathbf{A}\|_F^2 + \lambda \sum_{ij} C_{ij}^2 M_{ij}^{21}$$

Our unsupervised loss is a modified version of the one presented in \cite{roufosse2019unsupervised} and can be written as follows:
$$\mathcal{L}_{unsup} = \alpha_1 \mathcal{L}_{bij} + \alpha_2 \mathcal{L}_{orth},$$  
where :
\begin{itemize}
    \item  The bijectivity loss is formulated as follows: $\mathcal{L}_{bij} = \|C_{12}C_{21} - \mathbb{1}_r\|_F^2$. $\mathbb{1}_r$ is the identity matrix where only the first $r$ elements in the diagonal are equal to 1, $r$ is the estimated slope of the functional map under partiality, which we estimate using the approach proposed in \cite{Rodol2016}. Namely: $r = \max \{i | \Lambda_i^{2} < \max_{j = 1}^{k} \Lambda_j^1\}$. This loss promotes the bijectivity of the map, in the sense that transporting functions defined on the partial shape, using point-wise map, to the full shape and transporting them back should yield the same functions.
    
    \item The semi-orthogonality loss is formulated as: $\mathcal{L}_{orth} = \|C_{12}C_{12}^{\top} - \mathbb{1}_r\|_F^2 + \|C_{21}^{\top}C_{21} - \mathbb{1}_r\|_F^2$. This loss promotes the orthogonality of the functional map and thus local area preservation of the corresponding point-to-point map \cite{ovsjanikov2012functional}. Note that we are requiring only semi-orthogonality since the area preservation property holds only in the direction from partial to full shape.
    
    
\end{itemize}

In addition to these losses, and following existing literature \cite{roufosse2019unsupervised, SharmaO20}, it could be natural to use the commutativity with Laplacian loss: $\mathcal{L}_{comm} = \|C_{12}\Delta_1 - \Delta_2C_{12}\|_F^2 + \|C_{21}\Delta_2 - \Delta_1 C_{21}\|_F^2 $. This is not necessary in our case, however, as we already optimize for it during the construction of the functional map, in the Regularized FMap Module (see Section 3.4 of the main text).

To evaluate our unsupervised approach, we train our network using only the unsupervised loss with $\alpha_1 = \alpha_2 = 1$, by disabling our Cross Attention Refinement and Overlap Predictor modules, on the train set of the \textbf{CUTS} dataset, and evaluate it on the \textbf{PFARM} dataset. Results are reported in Figure \ref{fig:farm_unsup}. It can be seen that our unsupervised approach produces competitive results, as it outperforms all axiomatic methods, and gets on par with the \textit{supervised} learning-based baseline \cite{Donati2020}. Remarkably, our unsupervised approach generalizes across datasets and does not overfit to the underlying mesh structure, unlike the commonly used SHOT descriptors used in the axiomatic methods.


\begin{Figure}
  \includegraphics[width=1\linewidth]{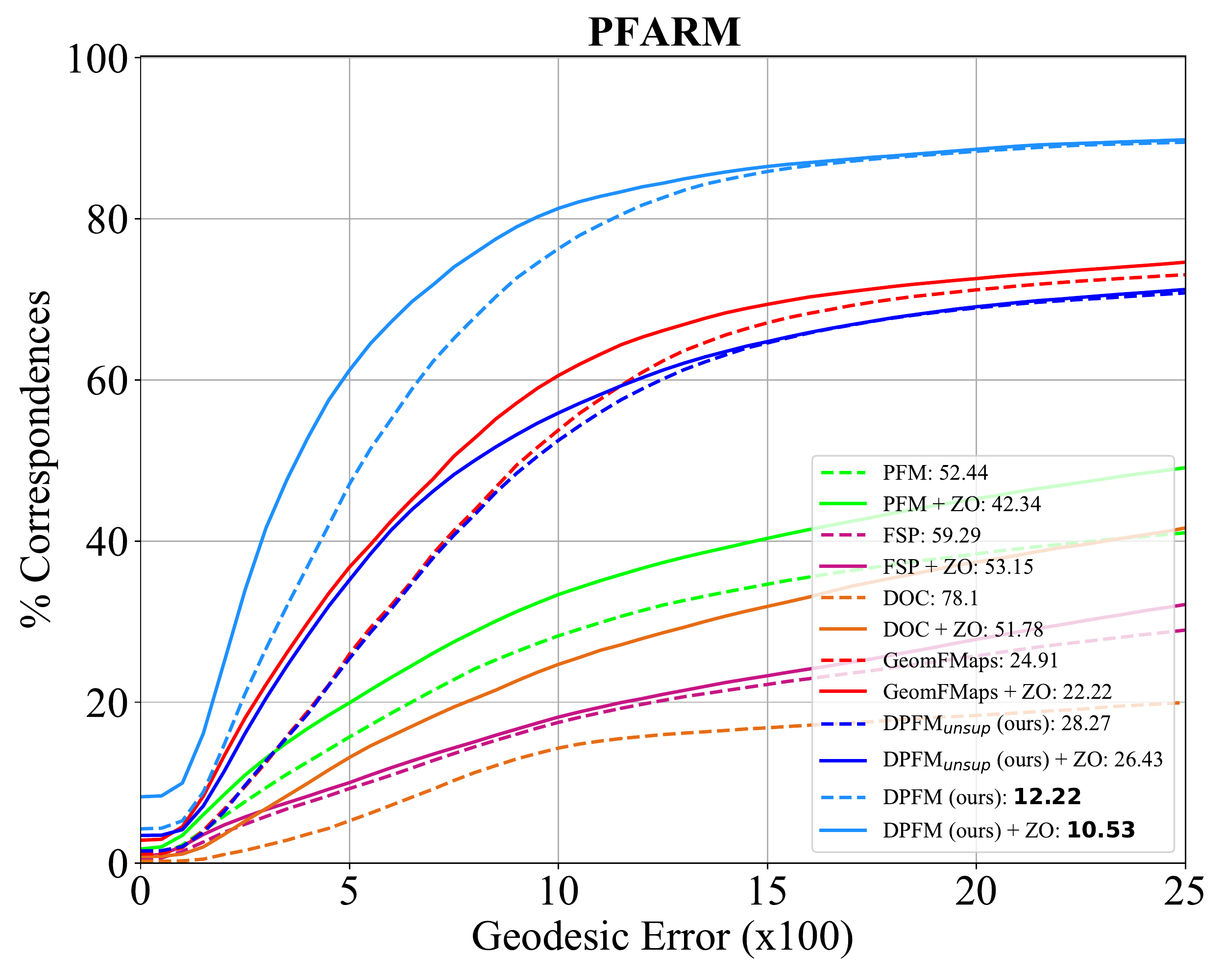}
\captionof{figure}{We demonstrate an unsupervised approach to our method on the \textbf{PFARM} dataset. Our method is competitive with our supervised learning-based baseline and significantly outperforms the axiomatic benchmarks.}
\label{fig:farm_unsup}
\end{Figure}

\section{Implementation and Network Training}
\label{sec:imp_details}
In Section 4 of the main text, we tested our network against multiple baselines, and in multiple settings.

In our experiments, we used two feature extractors: Diffusion-Net \cite{sharp2021diffusionnet} and SparseConvNet \cite{graham20183d}. For the former, we used the original implementation released by the authors \footnote{\url{https://github.com/nmwsharp/diffusion-net}}, our network is composed of four diffusion blocks of width 128, and outputs a final pointwise feature of size 128. For the latter, we used the implementation provided in the Minkowski Engine \footnote{\url{https://github.com/NVIDIA/MinkowskiEngine}}, our network has a Unet architecture \cite{RFB15a} of 4 blocks and outputs pointwise features of size 128.

Our Regularized FMap module in Section 3.4 of the main text aims to minimize the following energy:
$$C_{opt} = \argmin_{C} \|C \mathbf{A} - \mathbf{B}\|_F^2 + \lambda \sum_{ij} C_{ij}^2 M_{ij}$$
 We use the resolvent mask with the resolvent Laplacian parameter $\gamma = 0.5$, also, in all our experiments, we take $\lambda = 100$.

Our main supervised loss is composed of three terms and is written as follows: $\mathcal{L} = \lambda_1 \mathcal{L}_{spec} + \lambda_2 \mathcal{L}_{nce} + \lambda_3 \mathcal{L}_{over}$. For all our experiments, we took: $\lambda_1 = \lambda_2 = \lambda_3 = 1$, and the scaling parameter in $\mathcal{L}_{nce}$ is $\tau = 0.07$.

\mypara{Network training:} In all our experiments, we train the networks using an ADAM optimizer \cite{kingma2017adam} with an initial learning rate of 0.001. During training, we augment the training data on the fly by randomly rotating the input shapes around the up axis, applying random scaling in the range $[0.9, 1.1]$, and jitter the
position of each point by Gaussian noise with zero mean and 0.01 standard deviation, in order to make the network more robust, rotation invariant, and to avoid overfitting.

In order to recover the point-to-point map from the functional map, we used the standard nearest-neighbor method from the original functional map paper \cite{ovsjanikov2012functional}, and keep only the matches on the overlap region if necessary, using our predicted overlap mask.

As mentioned in the main manuscript, our code and data can be found online: \url{https://github.com/pvnieo/DPFM} to ensure full reproducibility of the results, and stimulate further research in this direction.

\mypara{Computational specifications}
All our experiments are executed using Pytorch \cite{NEURIPS2019_9015}, on a 64-bit machine, equipped with an Intel(R) Xeon(R) CPU E5-2630 v4 @ 2.20GHz and an RTX 2080 Ti Graphics Card.

\section{Datasets and Visualizations}
\label{sec:data_viz}

In Section 4.1 of the main document, we presented several datasets for training and evaluation. Namely, we used the SHREC16 Partial Correspondence Benchmark \cite{cosmo2016shrec}, which is a partial-to-full dataset. This dataset contains two subsets, \textbf{CUTS} and \textbf{HOLES}. \textbf{CUTS} is composed of 120 pairs for training, and 200 for testing, meanwhile, \textbf{HOLES} is composed of 80 pairs for training, and 200 for testing. Each partial shape is mapped to a null full shape which is a shape of the same class in a neutral pose. Some examples of this dataset are shown in Figure \ref{fig:dataset_vis} (top).

We also introduced a new dataset: \textbf{CP2P} which is aimed at evaluating partial-to-partial shape correspondence. In this dataset, partial shapes from the same class (either human or animals) are paired together. The overlap between the two shapes can range from 10\% to 90\%. Some examples of this dataset are shown in Figure \ref{fig:p2p_visual}.

Finally, we introduced \textbf{PFARM}, an extension of the recently introduced FARM partial dataset \cite{Kirgo2020}, which is a partial-to-full dataset, designed to test the robustness of partial shape matching methods to shapes that undergo near isometric deformations with a significant change of connectivity and sampling (see Figure \ref{fig:dataset_vis} - bottom). The partiality is imposed by segmenting and deleting random patches of shapes from the SHREC19 dataset \cite{marin2020farm}, and is composed of 27 different partial human shapes, that are all mapped to a full SMLP model \cite{SMPL_2015} of 6k vertices, resulting in 27 evaluation pairs. The resolution of each partial shape is around 10k vertices. It should be noted that because the size of this dataset is small, it was only used for evaluation, and never for training.

\begin{Figure}
  \includegraphics[width=1\linewidth]{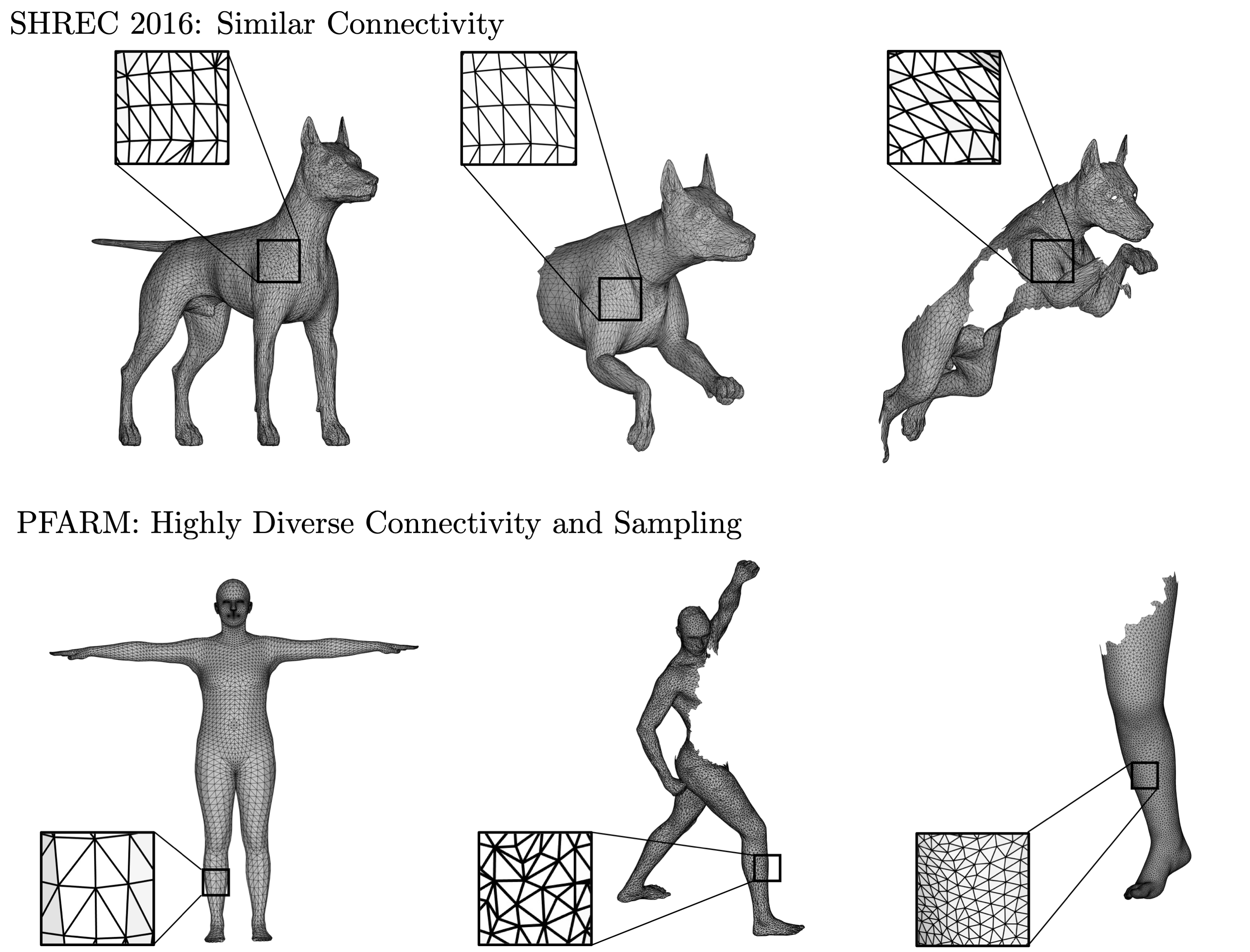}
\captionof{figure}{(top) Shapes from the SHREC16 Partiality Benchmark. Most of the shapes in this dataset have identical or very similar connectivity for both cuts and holes. (bottom) Shapes from the PFARM dataset that have highly diverse connectivity and sampling and provide a more challenging setting for dense partial shape correspondence.}
\label{fig:dataset_vis}
\end{Figure}

\begin{Figure}
\includegraphics[width=1\linewidth]{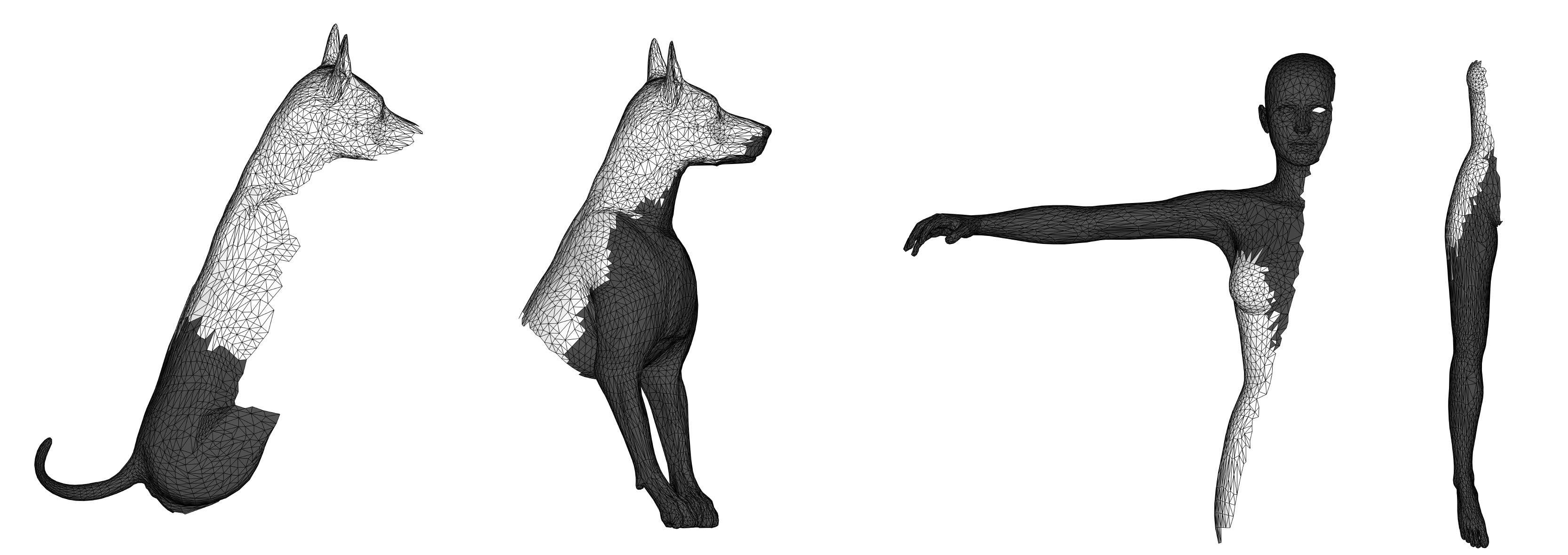}
\captionof{figure}{Example pairs from the {\bf CP2P} dataset with the non-corresponding regions indicated in grey (i.e., only the regions in white are expected to correspond). The dataset has some challenging pairs with a considerable amount of partiality similar to the second pair above.}
\label{fig:p2p_visual}
\end{Figure}

\section{Analysis of the Cross-Attention Refinement module}
\label{sec:refinement_module}
As was shown in Section 3 of the main manuscript, and corroborated by our theoretical analysis provided in Section A of the supplementary, the Cross Attention Refinement module enables the communication between features on the two shapes and thus allows the features on the overlap region to synchronize, while down-weighing the features outside the overlap. We evaluated this effect quantitatively on the entire \textbf{CP2P} dataset. Specifically, the percentage of points in the overlap region, with features whose $L_2$ norm is below a small threshold is \textbf{20\%} before the refinement and \textbf{23\%} after. Meanwhile, for the points on the non-overlapping region, this percentage changes from \textbf{34\%} before refinement to \textbf{83\%} after refinement. This demonstrates that the refinement module effectively processes the features to account for the points on the other shape, inside and outside the overlapping region, and consequently the overlap region prediction, as we observed a significant effect of the cross attention refinement on the quality of the overlap region prediction. Specifically, in the \textbf{CP2P} dataset, we measured the prediction accuracy to be \textbf{58\%} without the refinement module and \textbf{81\%} with our cross attention refinement processing.

\section{Quantitative evaluation}
\label{sec:quan_eva}
\begin{figure*}
\begin{subfigure}{.5\textwidth}
  \centering
  \includegraphics[width=1\linewidth]{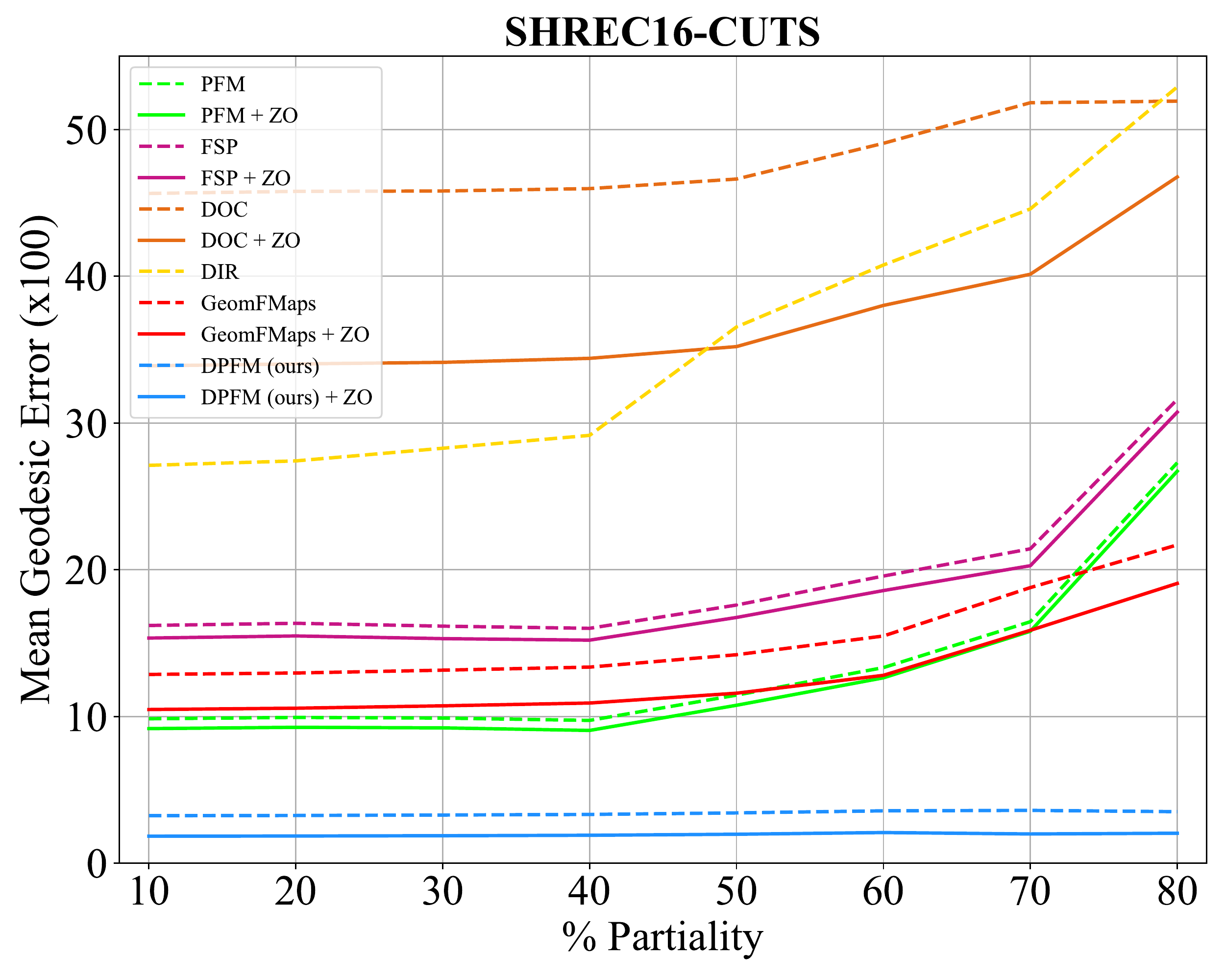}
  \caption{Cuts subsets}
  \label{fig:pc_partial_cuts}
\end{subfigure}%
\begin{subfigure}{.5\textwidth}
  \centering
  \includegraphics[width=1\linewidth]{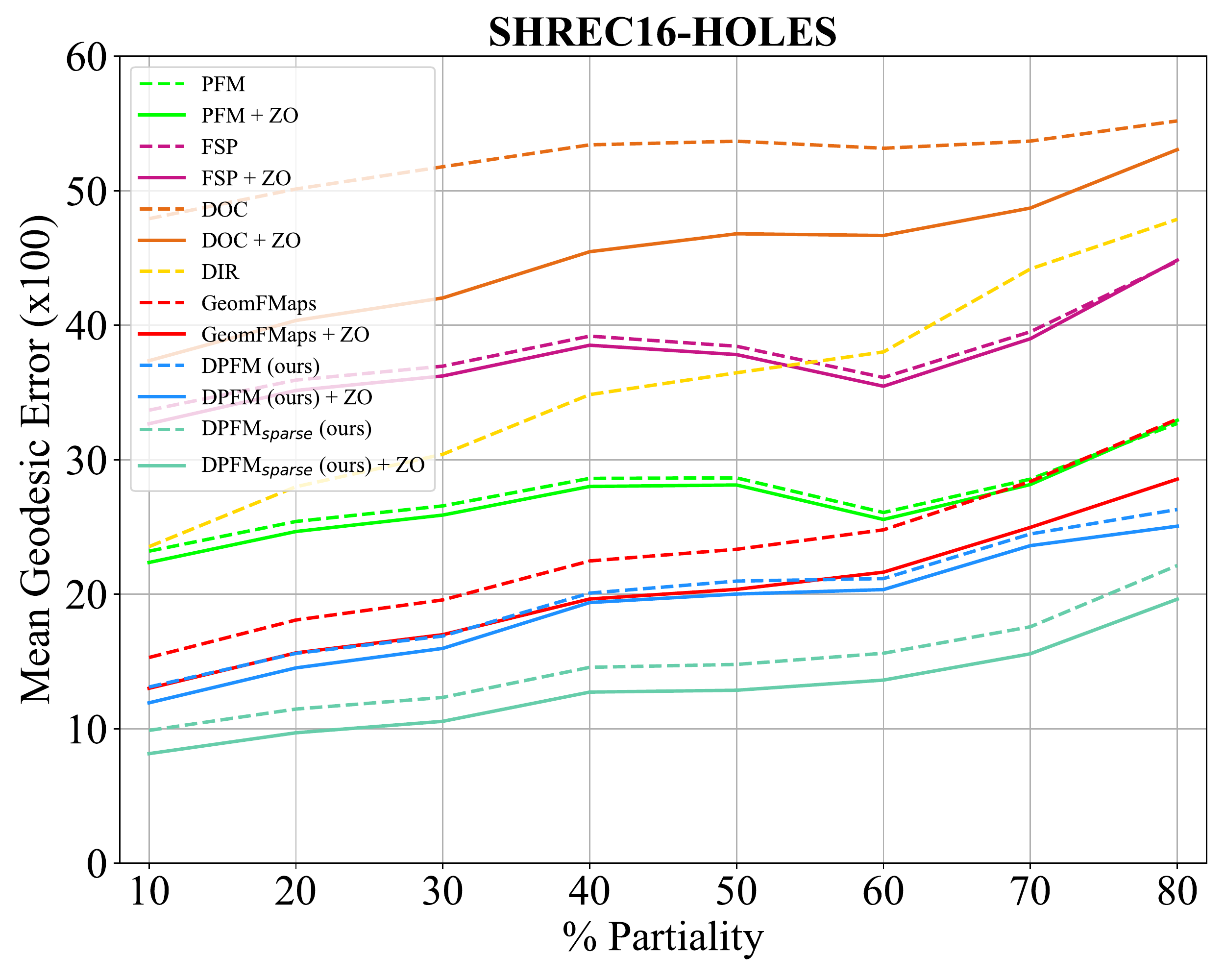}
  \caption{Holes subsets}
  \label{fig:pc_partial_holes}
\end{subfigure}
\caption{Performance of different methods in relation to the degree of partiality on the test set of SHREC'16 Partial Benchmark, both on cuts (left) and holes (right). Our method outperforms all the competing methods and achieves state-of-the-art results.}
\label{fig:pc_partial}
\end{figure*}

\begin{figure*}
\begin{subfigure}{.5\textwidth}
  \centering
  \includegraphics[width=1\linewidth]{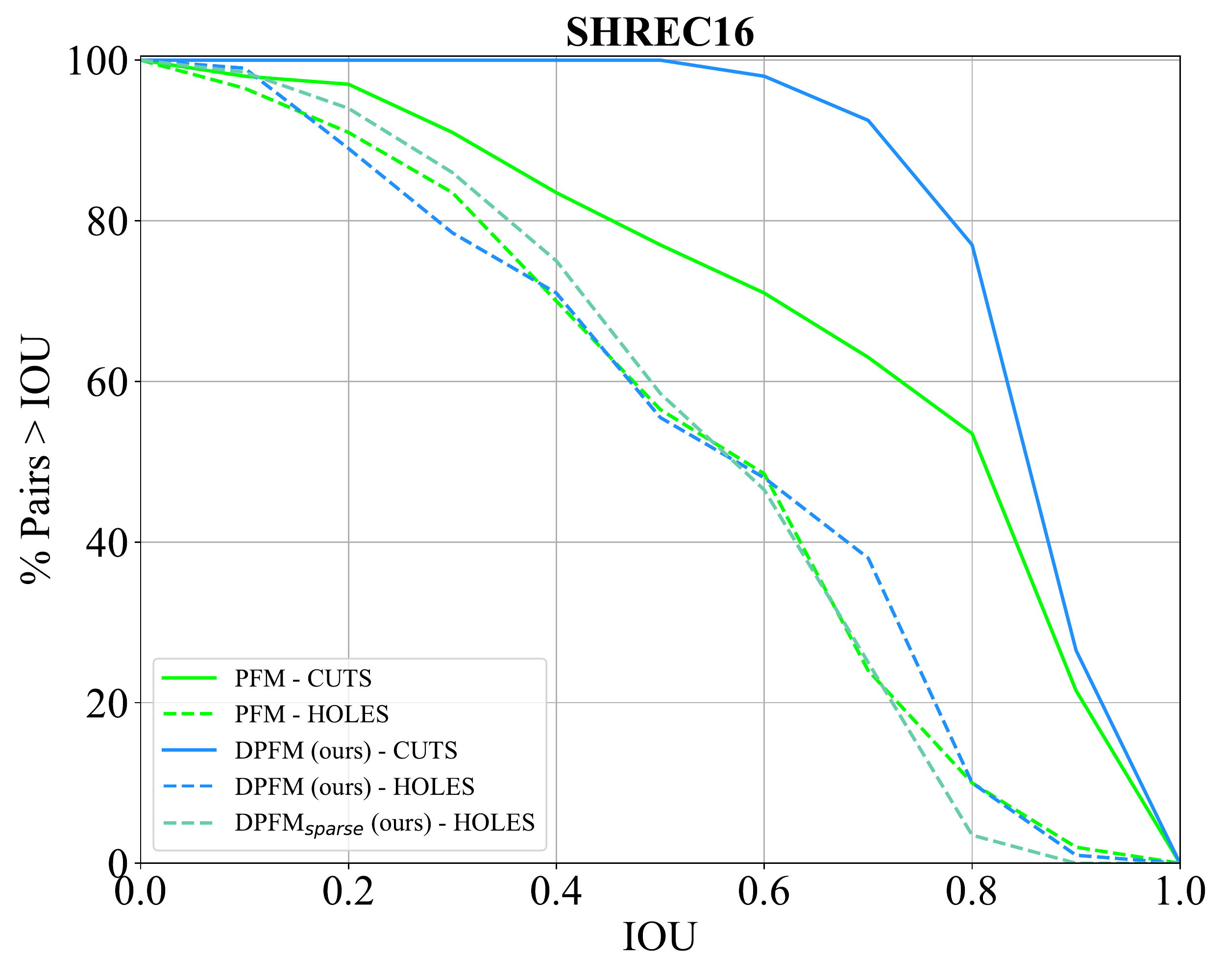}
  \caption{IOU for SHREC 16- cuts and holes}
  \label{fig:IOU_SHREC16}
\end{subfigure}%
\begin{subfigure}{.5\textwidth}
  \centering
  \includegraphics[width=1\linewidth]{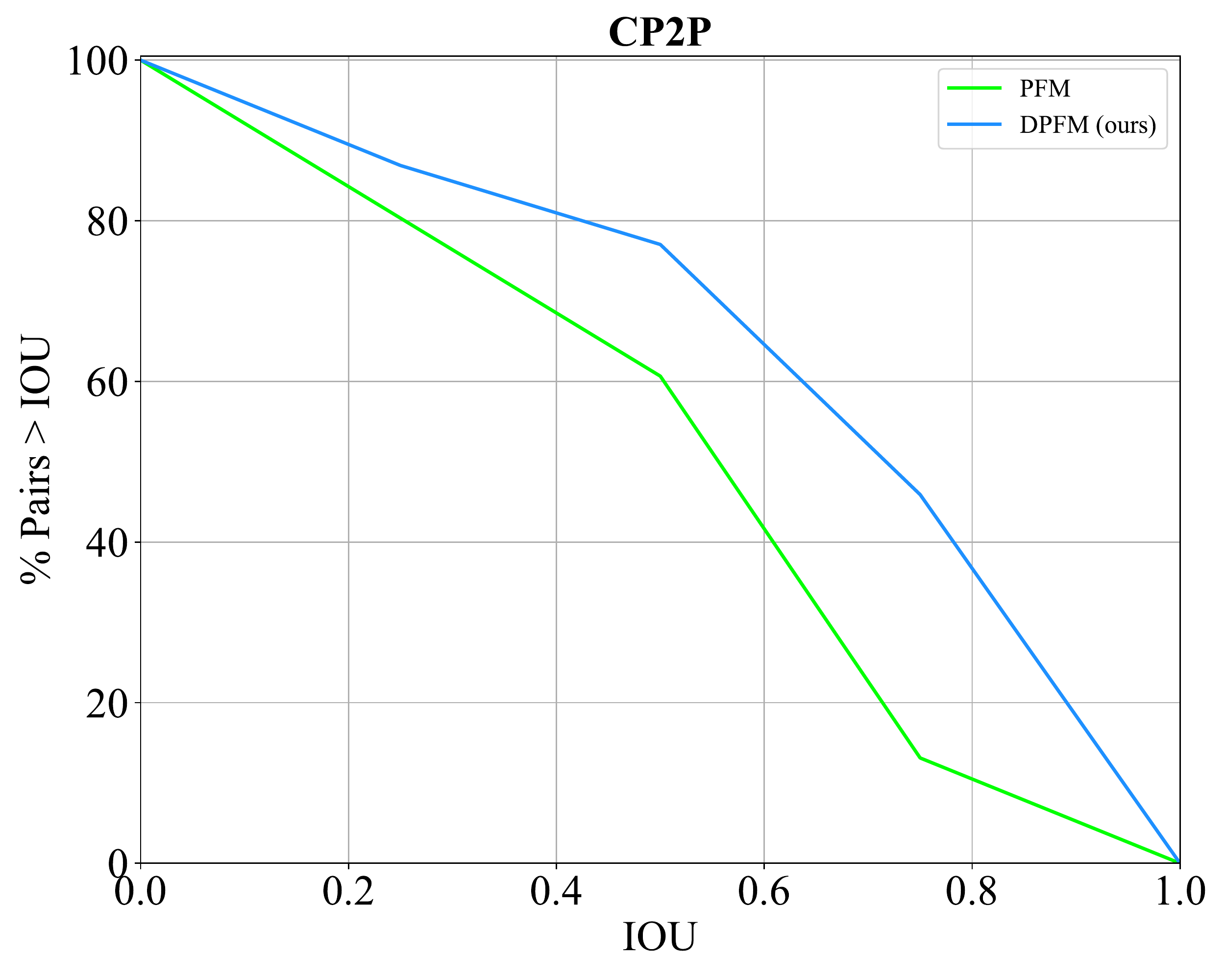}
  \caption{IOU: \textbf{CP2P}}
  \label{fig:IOU_p2p}
\end{subfigure}
\caption{Evaluation of region detection accuracy of our method vs PFM, on \textbf{CUTS}, \textbf{HOLES} and \textbf{CP2P}, by visualizing the evolution of the percentage of pairs having a certain IOU. It can be seen that our method achieves superior results, especially for \textbf{CUTS} and \textbf{CP2P}.}
\label{fig:iou}
\end{figure*}

\paragraph{Part To Full Shape Matching}
In Section 4.2 of the main manuscript, we show a comparison of our method with the baselines, on both \textbf{CUTS} and \textbf{HOLES} of the SHREC16 benchmark. In what follows, we provide some additional quantitative evaluation in Figures 
\ref{fig:pc_partial}, \ref{fig:IOU_SHREC16} and \ref{fig:IOU_p2p}. 


Specifically, we first show the average geodesic error as a function of the amount of partiality. 
We see in Figure \ref{fig:pc_partial} that our method has the lowest error curves compared to all baselines, obtaining state-of-the-art results. In addition, we see that our method stays significantly stable even in strong instances of partiality, especially for \textbf{CUTS}, which demonstrates the robustness of our method.
We additionally evaluate the accuracy of the predicted region, by plotting the intersection-over-union with the ground-truth region, for SHREC16, and compare it with the region predicted by PFM \cite{Rodol2016}. Figure \ref{fig:IOU_SHREC16} shows that our method gives significantly superior results than PFM, especially for cuts, where we obtain a very high IOU for a large number of pairs. 
\paragraph{Part To Part Shape Matching}
An evaluation of the correspondence accuracy of our method on the \textbf{CP2P} dataset was made in Section 4.4 of the main manuscript. Here we provide a quantitative evaluation of the region prediction ability. Since only our method, and the adaptation of PFM (recall that in order to predict the region using PFM on the \textbf{CP2P} dataset, we run it in both directions, which gives a prediction of the region on both the source and target shapes)  are capable of predicting the overlap region, we only evaluated these two methods. Figure \ref{fig:IOU_p2p} shows the evolution of the percentage of pairs having a certain IOU. It can be seen that our method outperforms PFM and obtains better results.


\section{Qualitative evaluation}
\label{sec:qual_eva}
In this section, we show some qualitative results of our method, as well as a comparison with the baselines.

\begin{figure*}[ht]
\begin{subfigure}{1\textwidth}
  \centering
  \includegraphics[width=1.0\linewidth]{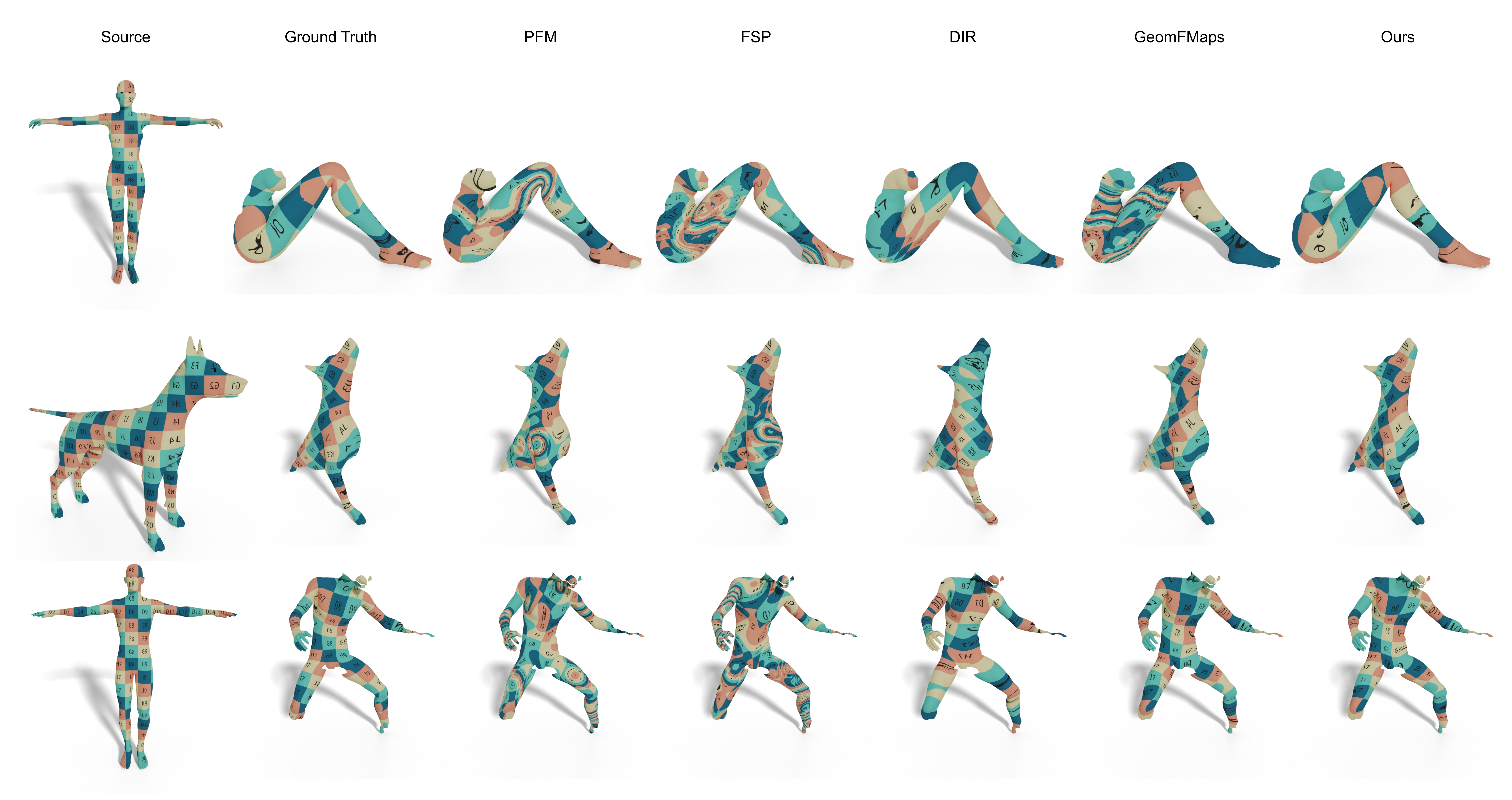}
  \caption{}
  \label{fig:quali_eva}
\end{subfigure}%

\begin{subfigure}{1\textwidth}
  \centering
  \includegraphics[width=1.0\linewidth]{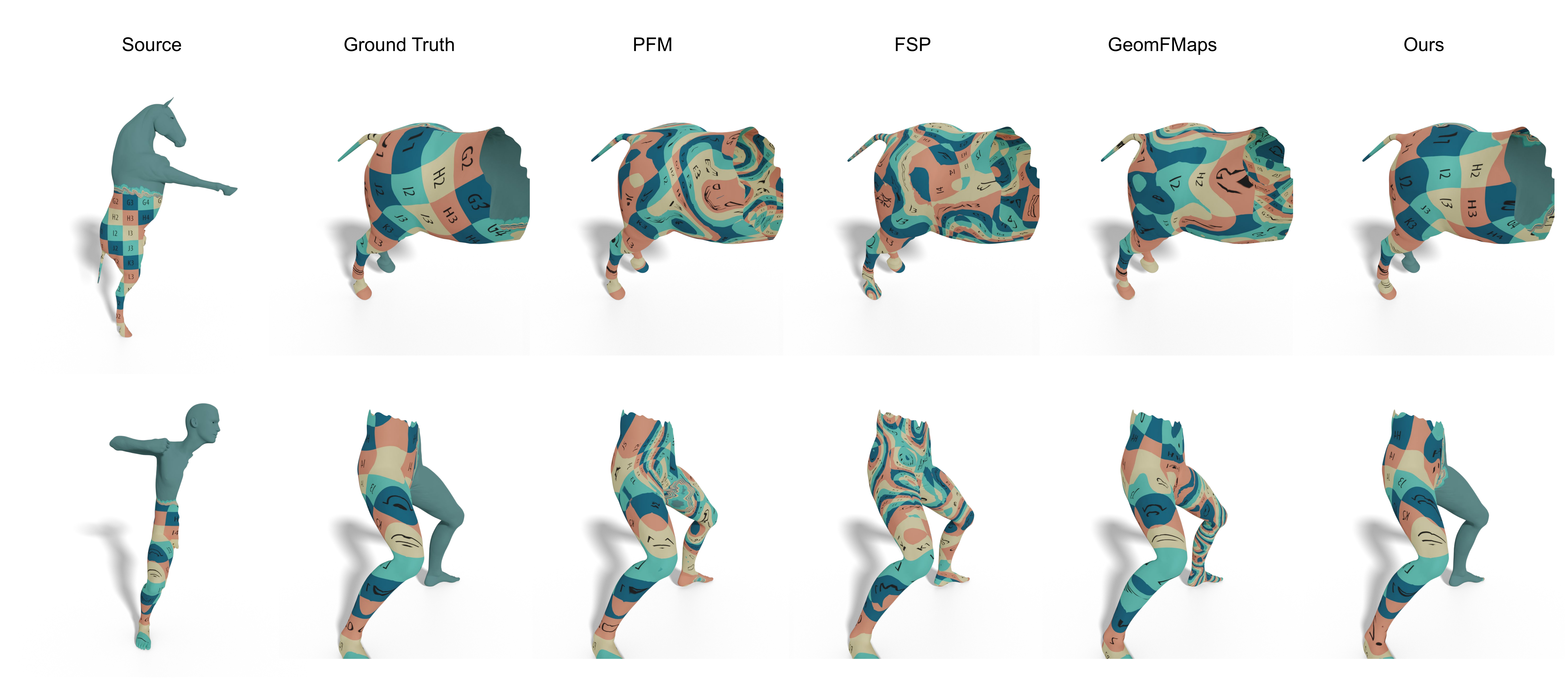}
  \caption{}
  \label{fig:quali_eva_p2p}
\end{subfigure}
\caption{(a) Qualitative comparison of our method and all the baselines, on \textbf{SHREC'16}, both on \textbf{CUTS} (top and middle) and \textbf{HOLES} (bottom). Correspondences are visualized by transferring a texture through the map. Our method yields visually plausible solutions on both cuts and holes, in both humans and animals. (b) Qualitative comparison of our method and all the baselines, on the \textbf{CP2P} dataset. Correspondences are visualized by transferring a texture through the map. Our method is the only one yielding visually plausible solutions and provides accurate region detection. The non-common regions are colored in green. \vspace{7mm}}
\label{fig:quali_eva_all}
\end{figure*}

Figure \ref{fig:quali_eva} visualizes the quality of the mapping using texture transfer on both \textbf{CUTS} and \textbf{HOLES}. It can be seen that our method achieves high-quality correspondences compared to the baselines.

Figure \ref{fig:quali_eva_p2p} shows the quality of the obtained map, and the region detected on \textbf{CP2P}, using texture transfer. It can be seen that our method is the only one that can accurately detect the overlap region, and provides accurate maps.

\begin{figure*}[t]
\begin{center}
\includegraphics[width=1.0\linewidth]{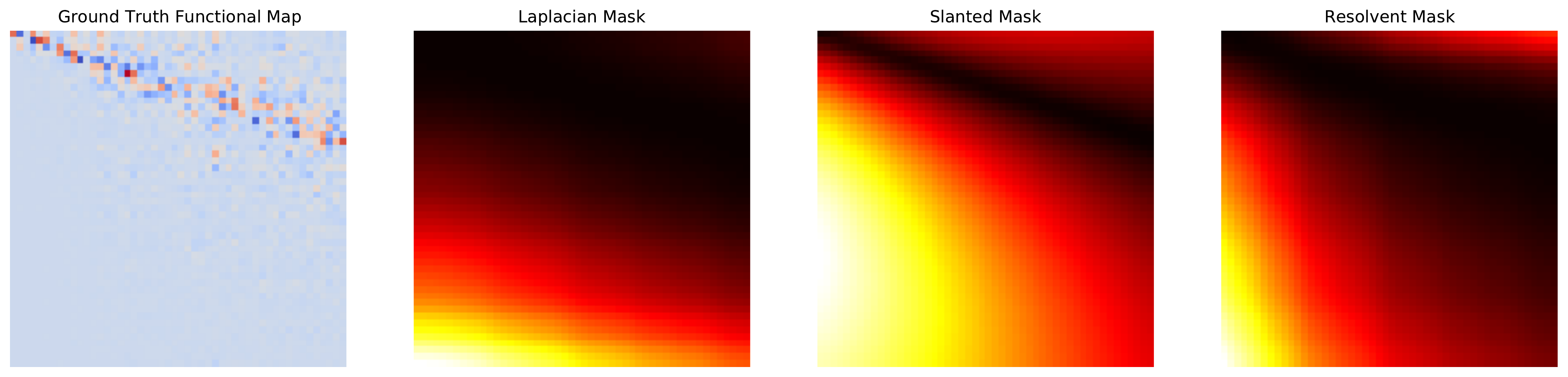}
\end{center}
\caption{Visualization of a ground-truth functional map, and the different masks tested. It appears that the resolvent mask is more adapted to the partial setting, as it follows the diagonal of the functional map, and is wider in the high frequencies. The slanted mask has a narrower diagonal, hence it penalizes the functional map more in the high frequencies, which harms the performance.}
\label{fig:masks}
\end{figure*}

\section{Ablation study}
\label{sec:abla}
In order to validate the different components of our approach, we consider two ablation studies, the first one concerns the choice of the mask used in the Regularized FMap module, and the second concerns the different terms in our proposed loss.

\paragraph{Mask Ablation}
In Section 3.4 of the main document, we proposed to use the resolvent mask for our Regularized FMap module. To confirm this choice, we train our network on the train set of \textbf{CUTS} and \textbf{HOLES}, and evaluate it on the test set of the same dataset. We conducted three experiments of this kind, the only variable is the used mask. We tested using the Laplacian Mask \cite{Donati2020}, the slanted mask \cite{Rodol2016} and the resolvent mask \cite{Ren2019}. Results are reported in Table~\ref{table:mask}. It can be seen that the resolvent mask helps to regularize the functional maps better, which leads to better results, especially in the challenging case of the \textbf{HOLES} dataset.

\begin{Table}
\begin{center}
\resizebox{\columnwidth}{!}{
\begin{tabular}{|l|c|c|c|}
\hline
\textbf{Mask} & \textbf{Laplacian Mask} & \textbf{Slanted Mask} & \textbf{Our Mask} \\
\hline\hline
CUTS & 3.5 & 4.27 & \textbf{3.2} \\
\hline
HOLES & 14.6 & 14.5 & \textbf{13.1} \\
\hline
\end{tabular}
}
\end{center}
\captionof{table}{The effects of using different masks in the Regularized FMap module. The reported mean geodesic errors are multiplied by 100 for clarity. The resolvent mask yields the best results, both in the case of cuts and holes, with significant improvement, especially in the latter case.}
\label{table:mask}
\end{Table}

To better illustrate this result, we visualize in Figure \ref{fig:masks} an example of a ground-truth functional map and the shape of the different masks. It can be seen that the Laplacian mask has a very large slanted region (in black), which provides poor regularization for the functional map. On the other hand, the slanted mask \cite{Rodol2016} is based on a heuristic, and from the shape of the mask, it can be seen that the latter promotes functional maps with a very narrow diagonal, which is not always good in the high frequencies. Finally, it can be seen that the resolvent mask \cite{Ren2019} follows the ground truth diagonal correctly, and the width of the latter changes as we move from low to high frequencies. 

Also, it should be noted that, unlike the Laplacian and the resolvent mask, the slanted mask requires to know exactly the direction of the slanted diagonal, information which is not available in the case of partial-to-partial, which limits the usability of this mask.

\paragraph{Loss Ablation}
Our proposed loss is composed of three terms. To validate the utility of each of them, we trained the sparse variant of our architecture and evaluated it on the test set of \textbf{HOLES}. We did four experiments, by training our network with all the losses, without the spectral loss $\mathcal{L}_{spec}$, without the NCE loss $\mathcal{L}_{nce}$, and finally without accuracy loss for the overlap module $\mathcal{L}_{over}$. It can be seen from Table \ref{table:loss_abla} that omitting any term of our loss leads to a significant drop in the performance. Observe also the importance of the spectral loss, as, without it, the network cannot converge, and no learning can be done. This suggests that the overlap prediction task benefits from the functional map correspondence learning. We verified this effect quantitatively on the \textbf{CP2P} dataset, where we observe that the accuracy of the predicted overlap region is 50\% without any functional map
correspondence learning and 81\% with it. Finally, we observe that the spectral loss alone is not enough for obtaining a good result, as it fails to provide important high-frequency information, especially in the challenging cases of holes, hence the need for the NCE loss.

\begin{Table}
\begin{center}
\begin{tabular}{cc}
\hline
\textbf{Ablation} & \textbf{Mean Error on HOLES}\\
\hline
no spectral loss & 28.3 \\
no NCE loss & 13.8 \\
no overlap loss & 10.0\\
Total loss & \textbf{9.3}\\
\hline
\end{tabular}
\end{center}
\captionof{table}{Ablation study of the different loss terms. The mean geodesic error is multiplied by 100 for clarity. Omitting any term of our proposed loss hurts the performance.}
\label{table:loss_abla}
\end{Table}

\end{multicols}

\end{document}